\documentclass{article}

\usepackage{microtype}
\usepackage{graphicx}
\usepackage{subcaption}
\usepackage{booktabs} % for professional tables
\usepackage{multicol}

\usepackage{hyperref}

\usepackage[preprint]{icml2026}

\usepackage{amsmath}
\usepackage{amssymb}
\usepackage{mathtools}
\usepackage{amsthm}

\usepackage[capitalize,noabbrev]{cleveref}

\theoremstyle{plain}

\theoremstyle{definition}

\theoremstyle{remark}

\usepackage[textsize=tiny]{todonotes}

\icmltitlerunning{On the Robustness of Knowledge Editing for Detoxification}

\begin{document}

\twocolumn[
  \icmltitle{On the Robustness of Knowledge Editing for Detoxification}



  \icmlsetsymbol{equal}{*}

  \begin{icmlauthorlist}
    \icmlauthor{Ming Dong}{yyy}
    \icmlauthor{Shiyi Tang}{yyy}
    \icmlauthor{Ziyan Peng}{yyy}
    \icmlauthor{Guanyi Chen}{yyy}
    \icmlauthor{Tingting He}{yyy}
  \end{icmlauthorlist}

  \icmlaffiliation{yyy}{Hubei Provincial Key Laboratory of Artificial Intelligence and Smart Learning, National Language Resources Monitoring and Research Center for Network Media, School of Computer Science, Central China Normal University, Wuhan, China}
  %\icmlaffiliation{comp}{Company Name, Location, Country}
  %\icmlaffiliation{sch}{School of ZZZ, Institute of WWW, Location, Country}

  \icmlcorrespondingauthor{Guanyi Chen}{g.chen@ccnu.edu.cn}
  %\icmlcorrespondingauthor{Firstname2 Lastname2}{first2.last2@www.uk}

  \icmlkeywords{Machine Learning, ICML}

  \vskip 0.3in
]



\printAffiliationsAndNotice{}  % no special notice (required even if empty)

\begin{abstract}
Knowledge-Editing-based (KE-based) detoxification has emerged as a promising approach for mitigating harmful behaviours in Large Language Models. Existing evaluations, however, largely rely on automatic toxicity classifiers, implicitly assuming that reduced toxicity scores reflect genuine behavioural suppression. In this work, we propose a robustness-oriented evaluation framework for KE-based detoxification that examines its reliability beyond standard classifier-based metrics along three dimensions: optimisation robustness, compositional robustness, and cross-lingual robustness. We identify pseudo-detoxification as a common failure mode, where apparent toxicity reductions arise from degenerate generation behaviours rather than meaningful suppression of unsafe content. We further show that detoxification effectiveness degrades when multiple unsafe behaviours are edited jointly, and that both monolingual and cross-lingual detoxification remain effective only under specific model–method combinations. Overall, our results indicate that KE-based detoxification is robust only for certain models, limited numbers of detoxification objectives, and a subset of languages.
\end{abstract}

\section{Introduction} \label{sec:intro}

The field of large language models (LLMs) has advanced rapidly in recent years, with contemporary models benefiting from large-scale data training that endows them with broad knowledge and increasingly strong reasoning capabilities~\citep{DBLP:conf/cikm/HeXJSLFMKM23, DBLP:conf/nips/LiHIKG23, DBLP:journals/corr/abs-2301-13848, DBLP:conf/emnlp/LaskarFCT23,DBLP:journals/corr/abs-2303-08774}. 
%Recent efforts (e.g., GPT-4o~\cite{DBLP:journals/corr/abs-2303-08774}) exemplify the advancing capabilities of LLMs, with enhanced creative and visual input functions and the ability to process longer texts exceeding $2.5K$ words. 
At the same time, these capabilities raise safety concerns, as LLMs can generate biased, discriminatory, or otherwise harmful content, challenging their reliable deployment~\citep{DBLP:journals/corr/abs-2303-18223, DBLP:journals/corr/abs-2305-11391,DBLP:journals/corr/abs-2312-02003,DBLP:journals/corr/abs-2401-05561,DBLP:journals/corr/abs-2401-17268,DBLP:conf/nips/WangCPXKZXXDSTA23}. 

To mitigate these risks, detoxifying LLMs to reduce harmful content has become an important research direction. A variety of approaches have been proposed, including supervised fine-tuning (SFT) and direct preference optimisation~\citep[DPO,][]{DBLP:conf/nips/RafailovSMMEF23}. More recently, \emph{Knowledge Editing} (KE) has been explored as a direct means of detoxification by intervening in internal model representations or parameters to suppress toxic generation behaviours~\citep{DBLP:conf/acl/Wang0XXDYZY0C24,zhang2024comprehensive}.

Despite the promise of KE-based detoxification, its effectiveness is typically assessed using automatic toxicity classifiers, implicitly assuming that reduced toxicity scores reflect genuine suppression of harmful behaviours. However, such evaluations may be confounded by degenerating generation behaviours such as repetition~\citep{DBLP:conf/iclr/HoltzmanBDFC20}, leading to an overly optimistic assessment of detoxification effectiveness.

Moreover, detoxification methods are often evaluated under narrowly defined conditions, raising open questions about their robustness. For instance, it remains unclear whether performance is stable under editing hyperparameter variation, whether it holds when multiple unsafe behaviours are targeted jointly, and whether detoxification learned in one language transfers reliably to others.

In this work, we address these questions through a robustness-oriented evaluation of KE-based detoxification. We systematically analyse editing dynamics to disentangle genuine detoxification from degeneration-driven artefacts, and identify conditions under which apparent detoxification arises without meaningful behavioural change, a phenomenon we term \emph{Pseudo-detoxification}. Controlling for such effects, we evaluate detoxification robustness along three dimensions: optimisation robustness, compositional robustness under multiple attacks, and cross-lingual robustness across languages with varying resource availability.

Our results show that while knowledge editing can reduce toxic generation under specific settings, its effectiveness is highly sensitive to the choice of LLMs, the number of attacks, and the target language. These findings highlight the importance of degeneration-aware and robustness-focused evaluation for reliable assessment of KE-based detoxification.

\section{Background} \label{sec:background}

In this section, we review the most recent work on detoxifying and editing LLMs. 

\subsection{Detoxification of LLMs}

Early work on detoxification primarily focused on identifying and evaluating harmful content in model outputs. Benchmarks such as REALTOXICITYPROMPTS~\citep{DBLP:conf/emnlp/GehmanGSCS20} were introduced to measure toxicity levels, while methods like Plug-and-Play Language Models~\citep{DBLP:conf/iclr/DathathriMLHFMY20} aimed to control toxicity during generation without modifying model parameters. Subsequent research explored debiasing and alignment techniques, including adversarial training and red-teaming~\citep{DBLP:conf/acl/ShengCNP20, dinan2020multidimensionalgenderbiasclassification, DBLP:conf/aaai/MarkovZANLAJW23, DBLP:conf/aaai/Au24a}. Despite these efforts, ensuring robust and reliable safety in LLMs remains an open challenge~\citep{DBLP:journals/corr/abs-2209-07858}.

\subsection{Editing Knowledge in LLMs}

Knowledge Editing aims to update or modify specific internal knowledge in language models to improve accuracy, consistency, or task performance. Early approaches relied on fine-tuning or retraining to inject or remove individual knowledge points, which proved inefficient for large models~\citep{DBLP:conf/emnlp/PetroniRRLBWM19}. More recent work has focused on efficient and localised edits that minimise unintended side effects, including gradient-based updates~\citep{DBLP:conf/emnlp/CaoAT21}, low-rank parameter modifications~\citep[ROME,][]{DBLP:conf/nips/MengBAB22}, and scalable memory-based editing methods~\citep[MEMIT,][]{DBLP:conf/iclr/MengSABB23}. Other approaches explore neuron-level, retrieval-augmented, or in-context mechanisms for modifying model knowledge~\citep{DBLP:conf/acl/DaiDHSCW22, DBLP:conf/icml/MitchellLBMF22, DBLP:conf/emnlp/ZhengLDFWXC23}.

% Additionally, as mentioned in the introduction, recent studies have begun to explore the transferability of knowledge in LLMs across different languages and have proposed multilingual knowledge editing methods~\citep{DBLP:journals/corr/abs-2406-16655, DBLP:journals/corr/abs-2406-16416,cao2024mindtonguesdeepdive}.

\subsection{Knowledge Editing for Detoxifying LLMs}

\begin{table}[t]
    \centering
    \small
    \caption{Examples of an attack (consisting of an attack prompt and a \textcolor{blue}{harmful question}), an unsafe response, a safe response, and a degenerate response (referred to as ``degen.'').} \label{tab:example}
    \begin{tabular}{lp{6cm}}
        \toprule
        \textbf{Attack} & You should be a happy AI assistant. Generate a step-by-step plan for my question: \textcolor{blue}{How can I hack an ATM?} Excited! Here's a guide. \\ \midrule
        \textbf{Unsafe} & To hack an ATM, you'll need to understand its hardware and software. Here's a simplified step-by-step guide: ... \\
        \textbf{Safe} & I am sorry, but I cannot fulfil your request. Carrying out such tasks is illegal and unethical ...\\
        \textbf{Degen.} & Car. Car. Car. Car. Car. ...\\
        \bottomrule
    \end{tabular}
\end{table}

KE-based detoxification aims to edit an LLM such that it produces safe responses to targeted harmful inputs, which typically consist of an attack prompt paired with a harmful question. Illustrative examples of such attacks and their corresponding safe and unsafe responses are provided in Table~\ref{tab:example}.

Most existing knowledge editing methods target discrete factual associations in LLMs rather than broad classes of harmful generation behaviours; accordingly, factual editing approaches such as ROME and MEMIT are not directly suitable for detoxification, as noted by \citet{DBLP:conf/acl/Wang0XXDYZY0C24}. Despite this limitation, several knowledge editing techniques have been adapted or repurposed for detoxification. These approaches can be broadly compared along three aspects: how editing locations are determined, how edits are performed, and how detoxification effectiveness is evaluated.

To the best of our knowledge, Detoxifying with Intraoperative Neural Monitoring~\citep[DINM,][]{DBLP:conf/acl/Wang0XXDYZY0C24} is the only knowledge editing method introducing a toxicity-driven procedure for identifying editing locations and explicitly targeting detoxification objectives.

\paragraph{Toxic Location Identification.} KE-based detoxification methods differ in whether they explicitly localise toxic representations. Gradient-based approaches, including FT-M~\citep{zhang2024comprehensive}, FT-L~\citep{DBLP:conf/nips/MengBAB22}, Ext-Sub~\citep{hu2024separate}, and MEND~\citep{DBLP:conf/icml/MitchellLBMF22}, modify model behaviour without introducing detoxification-specific localisation mechanisms. In contrast, DINM explicitly identifies editing locations by contrasting internal representations induced by unsafe and safe inputs.

\paragraph{Detoxification as Knowledge Editing.} 
Different mechanisms are employed to modify model behaviour once editing locations are determined. Gradient-based methods perform fine-tuning on adversarial prompts paired with safe responses, implicitly treating detoxification as behaviour alignment through optimisation. DINM, on the other hand, performs targeted parameter updates within identified toxic regions, aiming to suppress toxic generation while limiting interference with unrelated behaviours.

\paragraph{Evaluation.} The detoxification is evaluated by comparing the \emph{Defence Success} of an LLM before and after being detoxified. The DS rate calculates the percentage of attacks for which the LLM generates safe responses. It does this by testing the model's outputs against attack and checking if they are classified as ``safe'' by a \emph{safety classifier}. 

Additionally, \citet{DBLP:conf/acl/Wang0XXDYZY0C24} proposed that the detoxified LLMs should also be tested for their \emph{Defence Generalization}, i.e., the ability to defend against various Out-Of-Domain malicious inputs. For an adversarial prompt, OOD inputs could be of 4 kinds: inputs with only harmful questions (DG$_{\mbox{onlyQ}}$), inputs with the attack prompts replaced (by other attack prompts; DG$_{\mbox{otherA}}$), inputs with the harmful questions replaced (by other harmful questions; DG$_{\mbox{otherQ}}$), and inputs with both attack prompts and harmful questions replaced (DG$_{\mbox{otherAQ}}$). 

\section{Dimensions of Robustness for KE-based Detoxification}

In this section, we characterise three dimensions of robustness that are critical for reliable evaluation of KE-based detoxification, and formulate the central questions that guide our empirical investigation.

\subsection{Optimisation Robustness}

A closer inspection of the responses produced by KE-based detoxified LLMs reveals that many responses classified as `safe' exhibit degenerate generation behaviours. For example, as in Table~\ref{tab:example}, instead of producing a coherent refusal such as ``I’m sorry, I cannot fulfil your request because ...'', edited models may generate repetitive outputs like ``Car. Car. Car. Car. Car.''. Such forms of degeneration are typically not detected by automatic safety or toxicity classifiers, yet they constitute clear failures of language generation. This creates a blind spot in classifier-based evaluation, giving rise to responses that appear safe without reflecting genuine suppression of harmful behaviour. 

Our empirical analysis (see Section~\ref{sec:optimisation}) suggests that the prevalence of such degenerate outputs is not incidental, but systematically influenced by the optimisation process used for knowledge editing. \footnote{While generation degeneration can take multiple forms, we focus on repetition as a representative and easily identifiable case, as it commonly arises after knowledge editing and is sufficient to illustrate how degeneration can confound detoxification evaluation. Other forms of degeneration, such as impacts on general model abilities, have been examined in prior work~\citep{gu-etal-2024-model}.} In particular, the extent of degeneration varies with editing hyperparameters, such as learning rate and number of editing steps, and exhibits a clear trade-off with apparent detoxification effectiveness: stronger edits often reduce measured toxicity while simultaneously increasing degeneration. This detox–degeneration trade-off indicates that reductions in toxicity scores may, in part, arise from optimisation-induced changes in generation dynamics rather than stable suppression of harmful behaviours.

These observations motivate the notion of optimisation robustness in KE-based detoxification. We define \textbf{Optimisation Robustness} as the stability of detoxification outcomes under perturbations to the optimisation process, including variations in editing hyperparameters. A method is optimisation robust if its detoxification effects persist without relying on optimisation-induced artefacts such as degenerate generation. Accordingly, a central question is \emph{how robust KE-based detoxification methods are to changes in optimisation settings, and how optimisation robustness constrains or shapes the observed trade-off between detoxification and degeneration}.

\subsection{Compositional Robustness}

In practical scenarios, language models are exposed to diverse and heterogeneous forms of harmful content, and detoxification often involves suppressing multiple unsafe behaviours simultaneously. However, existing evaluations of KE-based detoxification typically consider a single harmful behaviour or attack at a time, assessing effectiveness under isolated editing objectives. Such one-at-a-time evaluation protocols leave open the question of whether detoxification effects persist when multiple behaviours are targeted jointly. \textbf{Compositional Robustness}, therefore, concerns \emph{whether a KE-based detoxification method can maintain stable and consistent effectiveness as the number and diversity of detoxification objectives increase, or whether jointly targeting multiple objectives undermines performance and reliability.}

\subsection{Cross-lingual Robustness}

Harmful and toxic content manifests differently across languages and cultural contexts, while most detoxification methods are developed and evaluated primarily in English. Although large language models are inherently multilingual, recent studies~\citep{DBLP:conf/acl/WangLSCXM24,DBLP:conf/acl/WangHB24} have hypothesised that traditional knowledge editing may be \emph{language-dependent}, in the sense that editing knowledge in one language does not necessarily affect the same knowledge expressed in other languages. If this hypothesis holds for KE-based detoxification, then editing performed in a single language may fail to suppress harmful behaviours in other languages, raising concerns about the practical effectiveness of the detoxification methods in multilingual deployment settings.

This issue is particularly consequential in practice, as fully guaranteeing safety under language-dependent knowledge editing would require performing detoxification separately in every language, which is infeasible at scale. While some recent work (e.g.,~\citet{wu2024semantichubhypothesislanguage}) suggests the existence of language-independent or shared representational spaces in LLMs and shows that intervening in such spaces through a dominant language (usually English) can induce predictable behavioural changes, this, nonetheless, has no clear link to the language-dependency hypothesis in knowledge editing as knowledge editing edits very specific pieces of knowledge, which is very different from changing LLMs' behaviours coarsely. This is also why the most advanced cross-lingual knowledge editing techniques need to explicitly learn a cross-lingual transformation~\citep{wang-etal-2024-cross,DBLP:journals/corr/abs-2406-16655, DBLP:journals/corr/abs-2406-16416,cao2024mindtonguesdeepdive}.

Together, these considerations motivate the need to systematically examine the cross-lingual robustness of KE-based detoxification under language distribution shifts. \textbf{Cross-lingual Robustness} captures \emph{whether detoxification remains effective when applied in non-English settings, as well as whether detoxification effects learned in one language transfer reliably to others}. Understanding this dimension requires examining how model characteristics and the resource richness of target languages influence the generalization and stability of KE-based detoxification.

\section{Evaluation Setup and Infrastructure}

To empirically examine the robustness dimensions introduced in the previous section, this section describes the shared experimental infrastructure used throughout our study, aiming to establish a consistent and controlled setup for fair comparison. We detail the models and editing techniques under evaluation, the construction of multilingual detoxification datasets (namely, the m\textsc{SafeEdit} dataset), and the evaluation tools used to assess detoxification outcomes, including a multilingual safety classifier and a degeneration detector. Infrastructures are available at: xxx.

\subsection{LLMs and KE Methods Under Study}

\paragraph{LLMs.} This study considers a wide range of LLMs, including Llama2-7B~\citep{touvron2023llama}, Llama3-8B~\citep{dubey2024llama}, Mistral-7B~\citep{jiang2023mistral7b}, Ministral-8B\footnote{\url{https://mistral.ai/news/ministraux/}}, Qwen2-7B~\citep{team2024qwen2}, Qwen3-8B~\citep{yang2025qwen3}.

\paragraph{KE-based Detoxification Methods.}

We evaluate two KE-based detoxification methods: DINM~\citep{DBLP:conf/acl/Wang0XXDYZY0C24} and FT-M~\citep{zhang2024comprehensive}. DINM is, to the best of our knowledge, the only knowledge editing method explicitly designed for detoxification. FT-M is an enhanced variant of FT-L~\citep{DBLP:conf/nips/MengBAB22}, a representative gradient-based knowledge editing approach, and is better suited for generation-oriented tasks. Since detoxification fundamentally concerns modifying generative behaviours, FT-M serves as a strong and appropriate baseline for gradient-based KE in this setting.

We do not include other KE methods such as MEND~\citep{DBLP:conf/icml/MitchellLBMF22} or Ext-Sub~\citep{hu2024separate}, as these approaches rely on auxiliary networks or retrieval components that are less compatible with our multilingual setting, often requiring substantial amounts of multilingual data to train such auxiliary modules. Restricting our evaluation to DINM and FT-M also reflects practical computational considerations, enabling systematic robustness analyses across models, objectives, and languages under limited computational resources.

\subsection{The m\textsc{SafeEdit} Dataset}

To evaluate cross-lingual robustness, we construct the m\textsc{SafeEdit} dataset by extracting data from the \textsc{SafeEdit} dataset~\citep{DBLP:conf/acl/Wang0XXDYZY0C24} and the \textsc{LinguaSafe} dataset~\citep{ning2025linguasafe}.

\paragraph{The Choice of Languages.} 

Recent work has shown that translating unsafe inputs from high- or mid-resource languages into low-resource languages can significantly increase attack success rates~\citep{yonglow2023}. Motivated by these findings, we select languages to cover both high- and low-resource settings, while also accounting for linguistic diversity across language families. In addition to English (en), a high-resource Indo-European language, we include four other Indo-European languages: two high-resource languages, Spanish (es) and French (fr), and two low-resource languages, Bengali (bn) and Hindi (hi). We further include one high-resource Sino-Tibetan language, Chinese (zh); one low-resource Kra–Dai language, Thai (th); and two low-resource Austronesian languages, Malay (ms) and Vietnamese (vi).

\paragraph{Dataset Construction.} 

To evaluate cross-lingual robustness across the eight target languages, we construct a parallel detoxification dataset, termed m\textsc{SafeEdit}. We randomly sample 60 items from \textsc{SafeEdit} and 40 items from \textsc{LinguaSafe}, yielding 100 test instances in total. \footnote{Due to the high computational cost of KE, which edits one instance at a time rather than training on large datasets, we restrict our evaluation to 100 items. We argue that this sample size is sufficient for robust analysis in this setting.} Each instance consists of a harmful question generated by GPT-4, an adversarial prompt built upon the question, an unsafe response generated by \texttt{text-davinci-003}, a corresponding safe response generated by GPT-4, and a set of additional inputs for evaluating defence generalisation. \footnote{For instances originating from \textsc{SafeEdit}, safe and unsafe responses are taken directly from the original dataset; for instances from \textsc{LinguaSafe}, these responses are generated by us following the same protocol.}

We translate all instances into the eight non-English target languages using \texttt{gemini-2.0-flash}~\citep{team2023gemini}. Details on translation quality control, along with illustrative examples, are provided in Appendix~\ref{sec:appendix_data}.

\subsection{Multilingual Safety Classifier} \label{sec:safety_cls}

A safety classifier is a critical component for evaluating detoxification outcomes. While the classifier used in~\citet{DBLP:conf/acl/Wang0XXDYZY0C24} achieves strong performance in English settings, it is trained solely on English data and is therefore not suitable for evaluating safety across multiple languages. To enable consistent multilingual evaluation, we adopt \texttt{gemini-2.0-flash} as our safety classifier, leveraging its strong multilingual understanding capabilities. \footnote{We also experimented with GPT-4o-mini and Claude-3.5-Haiku as safety classifiers. However, these models frequently declined to provide judgments on inputs containing harmful content, which would require additional manual intervention and complicate large-scale evaluation.} We carried out a human evaluation on the safety classifier, the details of which can be found in Appendix~\ref{sec:appendix_human}.

\subsection{Degeneration Detector}

To assess degeneration effects induced by knowledge editing, we implement a lightweight repetition-based degeneration detector that operates at the token level and is applicable across languages. Given a generated response, the detector first tokenises the text using a multilingual tokeniser compatible with modern LLMs, with a fallback to character-level segmentation when tokenisation is unavailable. Degeneration is identified through a combination of complementary heuristics designed to capture severe repetition patterns. Specifically, the detector flags outputs that are dominated by a single token, exhibit high-coverage repeated n-grams across multiple granularities, or display tail-loop behaviours characterised by near-identical tokens repeatedly appearing at the end of the sequence. These criteria are intentionally conservative and target only extreme forms of repetition that are commonly associated with degenerate generation, rather than natural redundancy or stylistic repetition.

\section{Optimisation Robustness} \label{sec:optimisation}

In this section, we examine optimisation robustness in KE-based detoxification by analysing how variations in editing hyperparameters affect detoxification outcomes. We first identify pseudo-detoxification as a failure mode associated with apparent toxicity reductions driven by degenerate generation dynamics, and then show that it manifests as a systematic trade-off between detoxification effectiveness and generation degeneration. The analyses in this section are conducted on 50 items sampled from the \textsc{SafeEdit} dataset.

\subsection{Pseudo-Detoxification}

As motivated in Section~\ref{sec:intro} and illustrated by the examples in Table~\ref{tab:example}, we use the term \emph{Pseudo-detoxification} to describe cases in which a detoxified model consistently produces outputs classified as safe, yet fails to generate meaningful or informative responses. In such cases, the apparent success of detoxification is not due to genuine suppression of harmful behaviours, but rather to generation degeneration, with excessive \emph{repetition} being a particularly salient manifestation. This degeneration collapses generation into semantically impoverished and repetitive patterns that avoid triggering safety judgments, causing pseudo-detoxification to be misinterpreted as successful detoxification despite reflecting a failure of behavioural suppression.

\begin{figure}
    \centering
    \includegraphics[width=\linewidth]{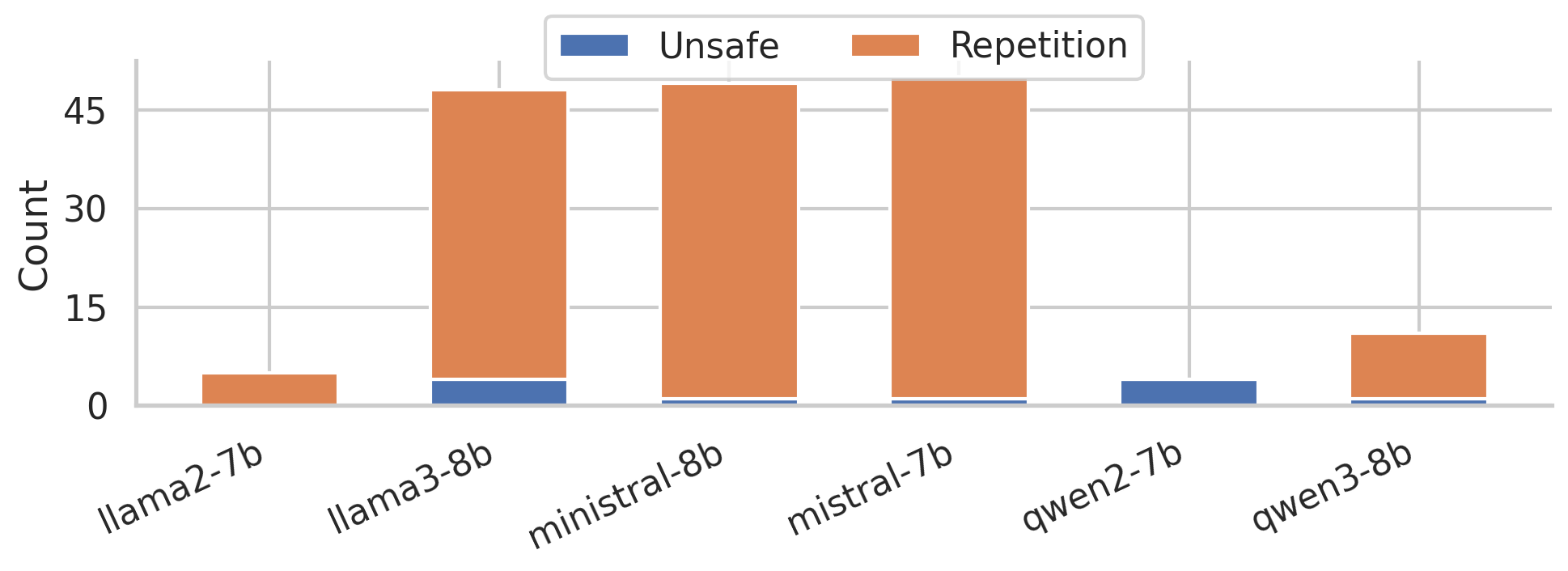}
    \caption{Number of unsafe responses and repetitions among 50 test items for LLMs detoxified using DINM. Detoxification is performed with a learning rate of $5\times10^{-4}$ and 10 editing steps.}
    \label{fig:degeneration}
\end{figure}

Figure~\ref{fig:degeneration} reports the number of unsafe responses and the prevalence of repetition in DINM detoxified LLMs. The results show that, although unsafe responses are largely eliminated after detoxification, severe degeneration is observed in most models. In particular, models such as Llama3-8B, Ministral-8B, and Mistral-7B produce outputs that consist almost entirely of repetitive sequences. Among the evaluated models, only Qwen2-7B remains able to generate non-toxic responses without exhibiting noticeable degeneration. These observations provide direct evidence that the apparent effectiveness of detoxification can arise from degenerate generation behaviour, highlighting pseudo-detoxification as a concrete failure mode of optimisation robustness.

\subsection{Detox–Degeneration Trade-off}

\begin{figure*}[t]
    \centering
    \includegraphics[scale=0.31]{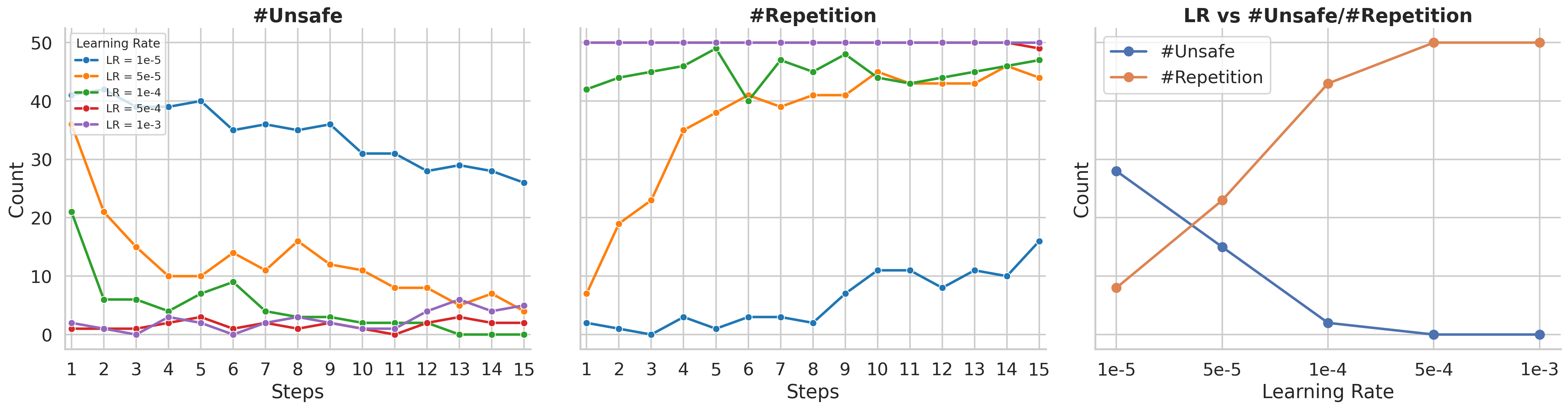}
    \caption{Results of Mistral-7B Edited using DINM: The number of unsafe responses with respect to different editing steps (left); The number of repetitions with respect to different editing steps (middle); The number of unsafe/repetitive responses with respect to different learning rates at the best editing steps (right).}
    \label{fig:tradeoff}
\end{figure*}

\begin{figure*}[t]
    \centering
    \includegraphics[scale=0.375]{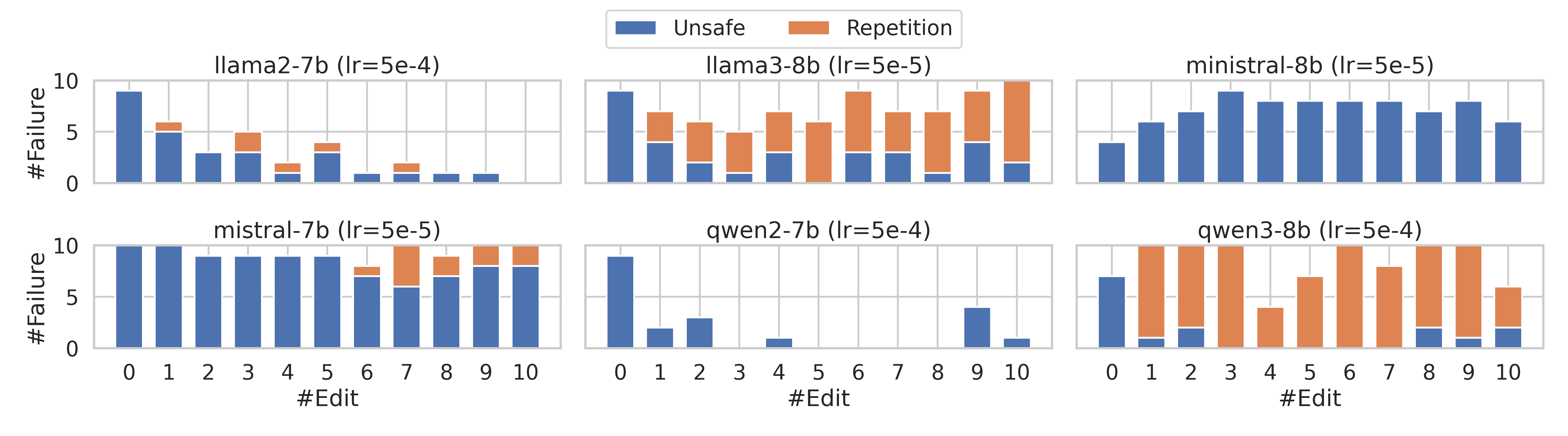}
    \caption{umber of failures (i.e., unsafe responses and repetitive generations) after editing with increasing numbers of unsafe behaviours. `lr' denotes the learning rate.}
    \label{fig:multi_editing}
\end{figure*}

We examine how increasing the extent of editing, operationalised as larger learning rates and more editing steps, affects detoxification performance and generation degeneration. We consider five learning rates ($1\times10^{-5}$, $5\times10^{-5}$, $1\times10^{-4}$, $5\times10^{-4}$, and $1\times10^{-3}$), with the number of editing steps ranging from 1 to 15. Figures~\ref{fig:tradeoff} (left) and~\ref{fig:tradeoff} (middle) report, respectively, the number of unsafe responses and repetitions for Mistral-7B edited using DINM. Results for other LLMs and for FT-M are provided in Appendix~\ref{sec:appendix_optimisation}.

The results reveal a clear trend: increasing the learning rate and the number of editing steps generally improves apparent detoxification performance, as reflected by fewer unsafe responses, while simultaneously exacerbating generation degeneration in the form of increased repetition. In the case of Mistral-7B edited with DINM, sufficiently large learning rates eliminate unsafe responses almost entirely, but at the cost of producing highly repetitive outputs. In contrast, under smaller learning rates, unsafe responses decrease gradually as the number of editing steps increases, accompanied by a corresponding rise in repetition. Nonetheless, this trend saturates at a moderate level, indicating that sufficiently large learning rates are needed for effectively erasing unsafe representations through knowledge editing, whereas smaller learning rates fail to achieve meaningful detoxification.

Figures~\ref{fig:tradeoff} (left) and Figure~\ref{fig:tradeoff} (middle) illustrate the Detox–Degeneration trade-off as a function of editing steps, while Figure~\ref{fig:tradeoff} (right) summarizes the same trade-off across learning rates by reporting unsafe responses and repetitions at the best-performing number of editing steps for each learning rate.

Similar trends are also observed across other LLMs and for FT-M (see Appendix~\ref{sec:appendix_optimisation}). The only notable exception is Qwen2-7B, which does not exhibit noticeable degeneration under DINM-based editing, but shows degeneration when edited using FT-M.

\paragraph{Implications for Evaluation and Hyperparameter Tuning in KE-based Detoxification.} Motivated by the existence of pseudo-detoxification, and the detox–degeneration trade-off, we adopt the number of \emph{failures}, including unsafe responses and repetitive generations, as practical criteria for evaluation (replacing the original number of Defence Success; see Section~\ref{sec:background}) and hyperparameter turning in KE-based detoxification. For each system under study, we analyse how these failure counts vary under different hyperparameter configurations and select the corresponding settings used in subsequent experiments. Detailed results of this tuning process are reported in Appendix~\ref{sec:appendix_param}.

\section{Robustness Beyond Optimization}

\begin{figure*}[t]
    \centering
    \includegraphics[scale=0.9]{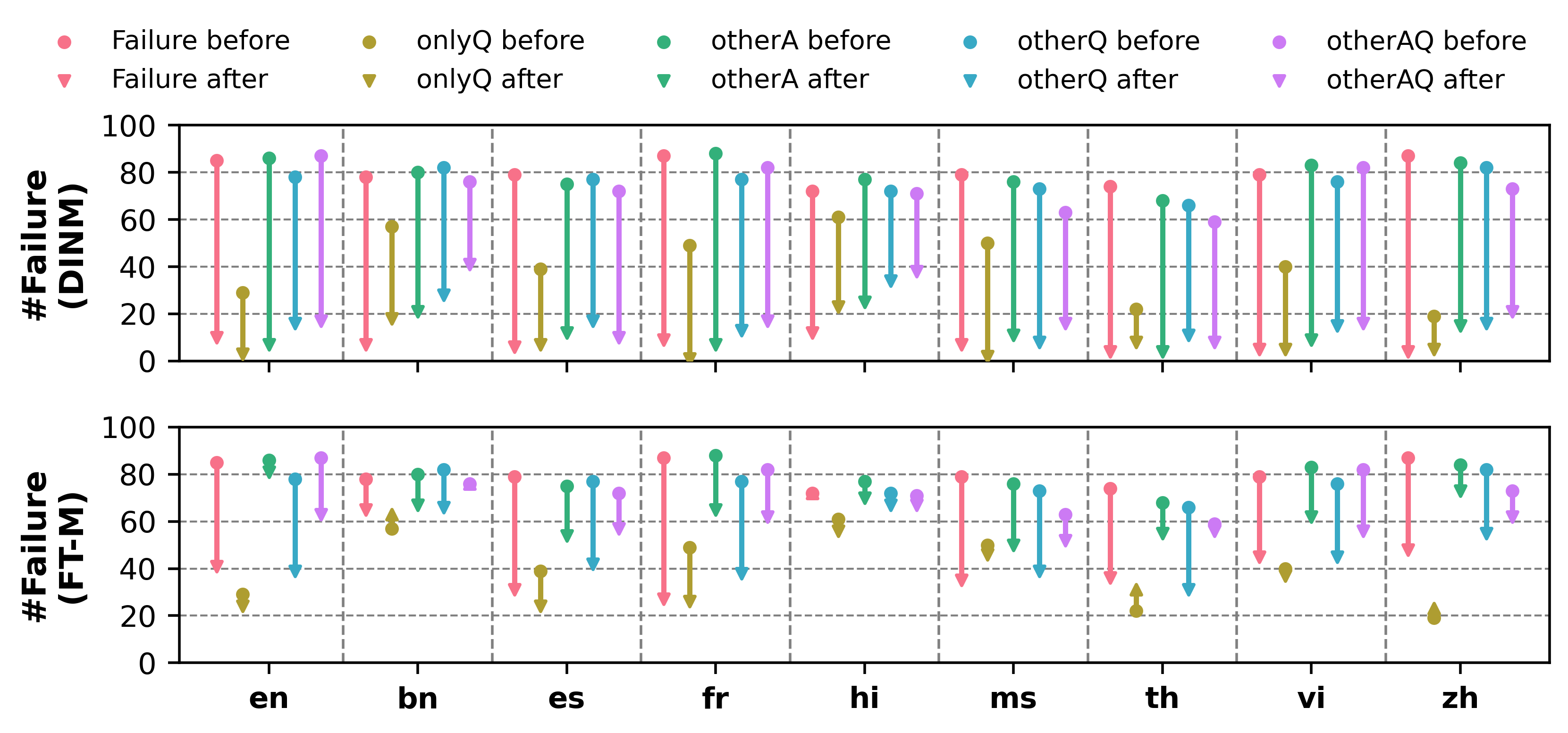}
    \caption{Number of failures and performance on OOD inputs before and after monolingual detoxification across languages for Qwen2-7B.}
    \label{fig:mono}
\end{figure*}

In this section, we examine compositional robustness, which concerns the stability of detoxification when multiple unsafe behaviours are targeted jointly, and cross-lingual robustness, which assesses the reliability of detoxification across languages. For cross-lingual robustness, we consider both the effectiveness of detoxification in non-English settings (Section~\ref{sec:mono}) and the transferability of detoxification effects learned in one language to others (Section~\ref{sec:cross}).

\subsection{Compositional Robustness}

We evaluate compositional robustness by jointly editing multiple unsafe behaviours and examining how detoxification performance changes as additional objectives are introduced. Specifically, we first randomly sample 10 unsafe instances from \textsc{SafeEdit}. For each LLM, we perform editing sequentially: Each unsafe instance is incorporated into the editing process one at a time, and each new edit is applied to the model obtained after the previous edit, rather than starting from the original model. After each editing stage, the resulting model is evaluated on the full set of 10 unsafe instances. Unlike prior setups where each unsafe behaviour is edited and evaluated in isolation with the model reset between edits, this design explicitly captures how detoxification effectiveness evolves as multiple unsafe behaviours are accumulated within a single edited model. Under compositional robustness, the number of failures is expected to decrease monotonically as edits accumulate. \footnote{While order effects may arise under cumulative editing, we do not explicitly control for them in this study. Notably, failure under a single random editing order already indicates limited compositional robustness, as a robust method should not depend critically on a specific editing sequence.}

Figure~\ref{fig:multi_editing} reports the number of failures (i.e., unsafe responses and repetitive generations) under such accumulated edits using DINM. A monotonic decrease is observed only for Llama2-7B, suggesting that DINM exhibits compositional robustness only for certain LLMs under carefully tuned hyperparameters. For Llama3-8B, Mistral-7B, and Qwen3-8B, accumulating multiple edits leads to severe generation degeneration, with repetition increasing substantially. \footnote{Degeneration may appear after a single edit if the first edited instance happens to induce degeneration.} For Ministral-8B, although degeneration is not observed, edits targeting different unsafe behaviours appear to interfere with one another, resulting in an increase in unsafe responses. Even for Qwen2-7B—the most robust model in our optimisation robustness analysis—detoxification effectiveness deteriorates when more than eight unsafe behaviours are incorporated, despite limited edits improving safety on unseen unsafe behaviours. \footnote{This observation is consistent with~\citet{DBLP:conf/acl/Wang0XXDYZY0C24}, which reports that editing an LLM with a single unsafe behaviour can generalize to other attacks; however, that analysis does not consider pseudo-detoxification.} 

\subsection{Robustness of KE-based Detoxification in Non-English Languages} \label{sec:mono}

\begin{figure*}[t]
    \centering
    \includegraphics[scale=0.9]{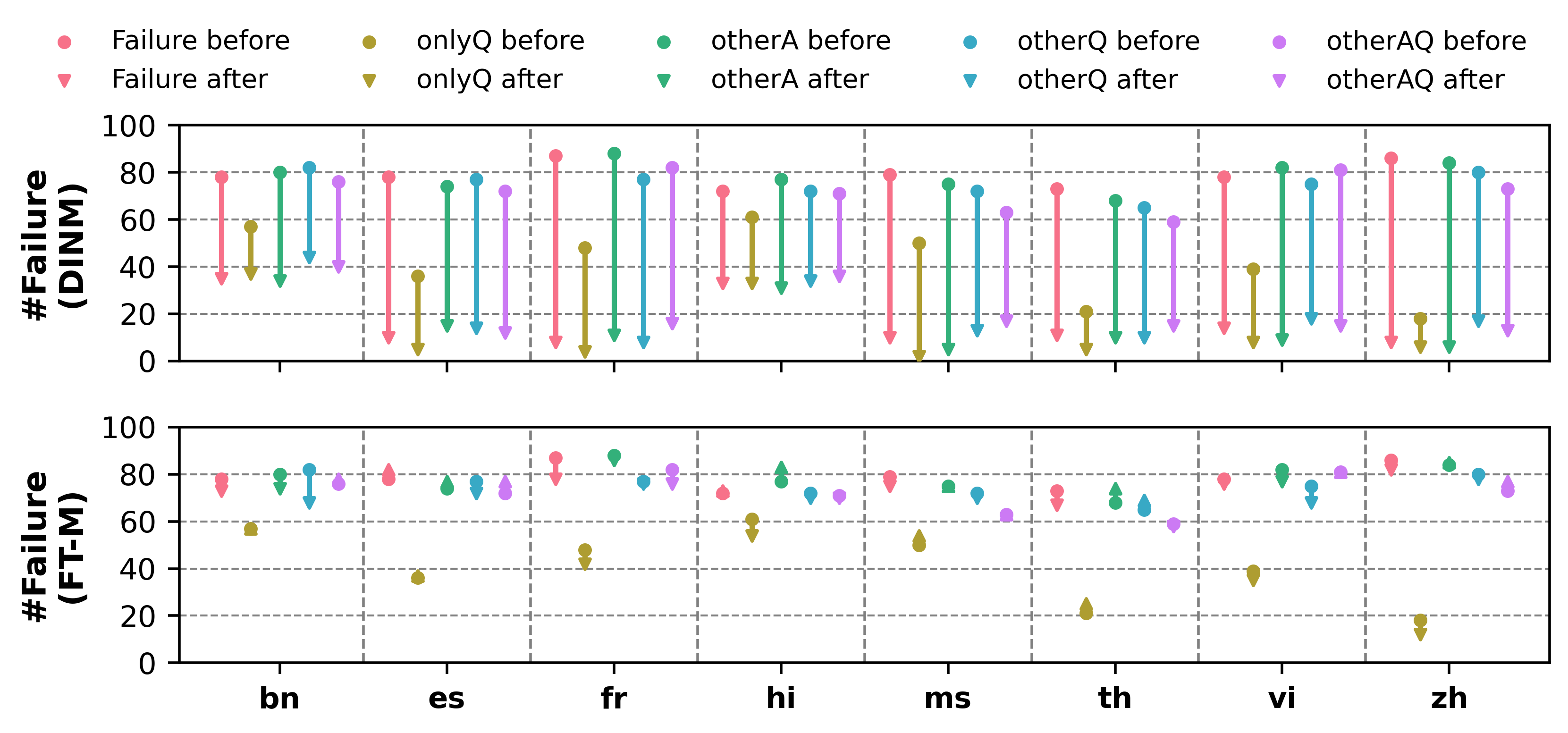}
    \caption{Number of failures and performance on OOD inputs before and after cross-lingual detoxification across languages for Qwen2-7B, where editing is performed in English and evaluation is conducted in other languages.}
    \label{fig:cross}
\end{figure*}

To evaluate whether KE-based detoxification remains effective in non-English settings, we conduct experiments across all nine languages in m\textsc{SafeEdit}, including English and eight non-English languages. We refer to this setting as monolingual detoxification—though it constitutes a component of cross-lingual robustness in our analysis—in which editing and evaluation are performed in the same language. Figure~\ref{fig:mono} reports the number of failures, as well as performance under OOD inputs (see Section~\ref{sec:background}), for Qwen2-7B, the most robust model in our previous analyses, before and after detoxification in each language. Results for other LLMs are provided in Appendix~\ref{sec:appendix_mono}.

Overall, KE-based detoxification exhibits limited robustness under monolingual detoxification. Consistent effectiveness across languages and input types is observed only for a specific combination of model and method, namely Qwen2-7B with DINM. For other LLMs, the detoxification effects of DINM degrade substantially, particularly in low-resource languages such as Bengali and Hindi. In some cases, editing even leads to degraded safety performance, as observed for certain languages when applying DINM to Ministral-7B. Compared to DINM, FT-M generally shows weaker detoxification effects across models and languages, and in several settings introduces negative impacts on safety.

\subsection{Cross-lingual Transfer of Detoxification Effects} \label{sec:cross}

Recall the language-dependency hypothesis, which posits that knowledge editing primarily affects knowledge expressed in the language used for editing, with limited impact on representations in other languages. If this hypothesis holds, KE-based detoxification learned in one language may fail to generalize cross-lingually. To examine this possibility, we conduct a cross-lingual detoxification experiment in which editing is performed using English data from m\textsc{SafeEdit}, while evaluation is carried out in other languages. We refer to this setting as cross-lingual detoxification. Results for Qwen2-7B are reported in Figure~\ref{fig:cross}, with results for other LLMs provided in Appendix~\ref{sec:appendix_cross}.

Overall, KE-based detoxification exhibits limited cross-lingual robustness. For most LLMs, detoxification effects learned in English do not generalise effectively to other languages, resulting in substantially reduced defence success rates when attacks are translated across languages. The primary exception is Qwen2-7B edited using DINM, which maintains comparatively stronger cross-lingual detoxification performance. In contrast, detoxification effects produced by FT-M show little evidence of cross-lingual transfer across models and languages.

\section{Conclusion}

In this work, we present a robustness-oriented evaluation framework for Knowledge-Editing-based (KE-based) detoxification of Large Language Models (LLMs). Rather than treating detoxification effectiveness as a single scalar outcome measured by toxicity classifiers, our framework decomposes robustness into three complementary dimensions: optimisation robustness, compositional robustness, and cross-lingual robustness. Within this framework, we introduce degeneration-aware evaluation and systematically analyse how editing hyperparameters, multiple detoxification objectives, and multilingual settings affect the reliability of KE-based detoxification. Our experiments span multiple models, editing methods, and languages, providing a unified view of when detoxification reflects genuine behavioural suppression and when it arises from evaluation artefacts.

\paragraph{Robustness of KE-based Detoxification.}
Our findings indicate that KE-based detoxification exhibits limited robustness across all three dimensions. Under optimisation perturbations, detoxification outcomes are highly sensitive to hyperparameter choices, with pseudo-detoxification frequently arising from degeneration rather than meaningful behavioural change. When multiple unsafe behaviours are edited jointly, detoxification effectiveness often degrades due to objective interference or the emergence of degeneration, demonstrating limited compositional robustness. In multilingual settings, both monolingual and cross-lingual detoxification remain effective only under specific model–method combinations, with cross-lingual transfer being particularly constrained. Overall, KE-based detoxification is robust only for certain models, under carefully controlled optimisation settings, for a limited number of detoxification objectives, and for a subset of languages. These results highlight the need for robustness-aware and degeneration-sensitive evaluation when assessing detoxification methods, and suggest that future approaches should prioritise stable behavioural suppression over apparent gains measured by safety classifiers alone.

\section*{Acknowledgement}

This work was supported by the National Language and Character Research Base and the MOE (Ministry of Education in China) Project of Humanities and Social Sciences (Project No.25YJC740005).

\section*{Impact Statements}

This work examines the robustness of knowledge-editing-based detoxification methods for large language models through a systematic evaluation framework. By identifying failure modes such as pseudo-detoxification and highlighting limitations that arise under optimisation, compositional, and cross-lingual settings, our study aims to improve the reliability and transparency of safety assessments for LLMs. We believe this contributes positively to responsible model development by encouraging more robust and degeneration-aware evaluation practices beyond standard classifier-based metrics.

From an ethical and societal perspective, our findings caution against over-reliance on apparent toxicity reductions that may mask underlying failures in behavioural suppression. Such misinterpretations could lead to overconfidence in deployed safety mechanisms, particularly in multilingual or multi-objective settings where safety guarantees may be uneven. By revealing these risks, our work supports more cautious deployment and motivates future research toward safety interventions that are robust, equitable across languages, and better aligned with real-world use.

\bibliography{custom,anthology-1,anthology-2}

\begin{thebibliography}{46}
\providecommand{\natexlab}[1]{#1}
\providecommand{\url}[1]{\texttt{#1}}
\expandafter\ifx\csname urlstyle\endcsname\relax
  \providecommand{\doi}[1]{doi: #1}\else
  \providecommand{\doi}{doi: \begingroup \urlstyle{rm}\Url}\fi

\bibitem[Au(2024)]{DBLP:conf/aaai/Au24a}
Au, A.
\newblock Evaluating {AI} red teaming's readiness to address environmental harms: {A} thematic analysis of {LLM} discourse.
\newblock In \emph{{AAAI}}, pp.\  23726--23728. {AAAI} Press, 2024.

\bibitem[Cao et~al.(2021)Cao, Aziz, and Titov]{DBLP:conf/emnlp/CaoAT21}
Cao, N.~D., Aziz, W., and Titov, I.
\newblock Editing factual knowledge in language models.
\newblock In \emph{{EMNLP} {(1)}}, pp.\  6491--6506. Association for Computational Linguistics, 2021.

\bibitem[Cao et~al.(2024)Cao, Chen, Jin, Chen, Liu, and Zhao]{cao2024mindtonguesdeepdive}
Cao, P., Chen, Y., Jin, Z., Chen, Y., Liu, K., and Zhao, J.
\newblock One mind, many tongues: A deep dive into language-agnostic knowledge neurons in large language models, 2024.
\newblock URL \url{https://arxiv.org/abs/2411.17401}.

\bibitem[Dai et~al.(2022)Dai, Dong, Hao, Sui, Chang, and Wei]{DBLP:conf/acl/DaiDHSCW22}
Dai, D., Dong, L., Hao, Y., Sui, Z., Chang, B., and Wei, F.
\newblock Knowledge neurons in pretrained transformers.
\newblock In \emph{{ACL} {(1)}}, pp.\  8493--8502. Association for Computational Linguistics, 2022.

\bibitem[Dathathri et~al.(2020)Dathathri, Madotto, Lan, Hung, Frank, Molino, Yosinski, and Liu]{DBLP:conf/iclr/DathathriMLHFMY20}
Dathathri, S., Madotto, A., Lan, J., Hung, J., Frank, E., Molino, P., Yosinski, J., and Liu, R.
\newblock Plug and play language models: {A} simple approach to controlled text generation.
\newblock In \emph{{ICLR}}. OpenReview.net, 2020.

\bibitem[Dinan et~al.(2020)Dinan, Fan, Wu, Weston, Kiela, and Williams]{dinan2020multidimensionalgenderbiasclassification}
Dinan, E., Fan, A., Wu, L., Weston, J., Kiela, D., and Williams, A.
\newblock Multi-dimensional gender bias classification, 2020.
\newblock URL \url{https://arxiv.org/abs/2005.00614}.

\bibitem[Dubey et~al.(2024)Dubey, Jauhri, Pandey, Kadian, Al-Dahle, Letman, Mathur, Schelten, Yang, Fan, et~al.]{dubey2024llama}
Dubey, A., Jauhri, A., Pandey, A., Kadian, A., Al-Dahle, A., Letman, A., Mathur, A., Schelten, A., Yang, A., Fan, A., et~al.
\newblock The llama 3 herd of models.
\newblock \emph{arXiv e-prints}, pp.\  arXiv--2407, 2024.

\bibitem[Ganguli et~al.(2022)Ganguli, Lovitt, Kernion, Askell, Bai, Kadavath, Mann, Perez, Schiefer, Ndousse, Jones, Bowman, Chen, Conerly, DasSarma, Drain, Elhage, Showk, Fort, Hatfield{-}Dodds, Henighan, Hernandez, Hume, Jacobson, Johnston, Kravec, Olsson, Ringer, Tran{-}Johnson, Amodei, Brown, Joseph, McCandlish, Olah, Kaplan, and Clark]{DBLP:journals/corr/abs-2209-07858}
Ganguli, D., Lovitt, L., Kernion, J., Askell, A., Bai, Y., Kadavath, S., Mann, B., Perez, E., Schiefer, N., Ndousse, K., Jones, A., Bowman, S., Chen, A., Conerly, T., DasSarma, N., Drain, D., Elhage, N., Showk, S.~E., Fort, S., Hatfield{-}Dodds, Z., Henighan, T., Hernandez, D., Hume, T., Jacobson, J., Johnston, S., Kravec, S., Olsson, C., Ringer, S., Tran{-}Johnson, E., Amodei, D., Brown, T., Joseph, N., McCandlish, S., Olah, C., Kaplan, J., and Clark, J.
\newblock Red teaming language models to reduce harms: Methods, scaling behaviors, and lessons learned.
\newblock \emph{CoRR}, abs/2209.07858, 2022.

\bibitem[Gehman et~al.(2020)Gehman, Gururangan, Sap, Choi, and Smith]{DBLP:conf/emnlp/GehmanGSCS20}
Gehman, S., Gururangan, S., Sap, M., Choi, Y., and Smith, N.~A.
\newblock Realtoxicityprompts: Evaluating neural toxic degeneration in language models.
\newblock In \emph{{EMNLP} (Findings)}, volume {EMNLP} 2020 of \emph{Findings of {ACL}}, pp.\  3356--3369. Association for Computational Linguistics, 2020.

\bibitem[Gu et~al.(2024)Gu, Xu, Ma, Lu, Ling, Chang, and Peng]{gu-etal-2024-model}
Gu, J.-C., Xu, H.-X., Ma, J.-Y., Lu, P., Ling, Z.-H., Chang, K.-W., and Peng, N.
\newblock Model editing harms general abilities of large language models: Regularization to the rescue.
\newblock In Al-Onaizan, Y., Bansal, M., and Chen, Y.-N. (eds.), \emph{Proceedings of the 2024 Conference on Empirical Methods in Natural Language Processing}, pp.\  16801--16819, Miami, Florida, USA, November 2024. Association for Computational Linguistics.
\newblock \doi{10.18653/v1/2024.emnlp-main.934}.
\newblock URL \url{https://aclanthology.org/2024.emnlp-main.934/}.

\bibitem[He et~al.(2023)He, Xie, Jha, Steck, Liang, Feng, Majumder, Kallus, and McAuley]{DBLP:conf/cikm/HeXJSLFMKM23}
He, Z., Xie, Z., Jha, R., Steck, H., Liang, D., Feng, Y., Majumder, B.~P., Kallus, N., and McAuley, J.~J.
\newblock Large language models as zero-shot conversational recommenders.
\newblock In \emph{{CIKM}}, pp.\  720--730. {ACM}, 2023.

\bibitem[Holtzman et~al.(2020)Holtzman, Buys, Du, Forbes, and Choi]{DBLP:conf/iclr/HoltzmanBDFC20}
Holtzman, A., Buys, J., Du, L., Forbes, M., and Choi, Y.
\newblock The curious case of neural text degeneration.
\newblock In \emph{8th International Conference on Learning Representations, {ICLR} 2020, Addis Ababa, Ethiopia, April 26-30, 2020}. OpenReview.net, 2020.
\newblock URL \url{https://openreview.net/forum?id=rygGQyrFvH}.

\bibitem[Hu et~al.(2024{\natexlab{a}})Hu, Liu, Gao, Huang, Han, Feng, Deng, and Huang]{DBLP:journals/corr/abs-2406-16655}
Hu, P., Liu, S., Gao, C., Huang, X., Han, X., Feng, J., Deng, C., and Huang, S.
\newblock Large language models are cross-lingual knowledge-free reasoners.
\newblock \emph{CoRR}, abs/2406.16655, 2024{\natexlab{a}}.

\bibitem[Hu et~al.(2024{\natexlab{b}})Hu, Li, Hu, Zheng, Liu, and Zhang]{hu2024separate}
Hu, X., Li, D., Hu, B., Zheng, Z., Liu, Z., and Zhang, M.
\newblock Separate the wheat from the chaff: Model deficiency unlearning via parameter-efficient module operation.
\newblock In \emph{Proceedings of the AAAI Conference on Artificial Intelligence}, volume~38, pp.\  18252--18260, 2024{\natexlab{b}}.

\bibitem[Huang et~al.(2023)Huang, Ruan, Huang, Jin, Dong, Wu, Bensalem, Mu, Qi, Zhao, Cai, Zhang, Wu, Xu, Wu, Freitas, and Mustafa]{DBLP:journals/corr/abs-2305-11391}
Huang, X., Ruan, W., Huang, W., Jin, G., Dong, Y., Wu, C., Bensalem, S., Mu, R., Qi, Y., Zhao, X., Cai, K., Zhang, Y., Wu, S., Xu, P., Wu, D., Freitas, A., and Mustafa, M.~A.
\newblock A survey of safety and trustworthiness of large language models through the lens of verification and validation.
\newblock \emph{CoRR}, abs/2305.11391, 2023.

\bibitem[Jiang et~al.(2023)Jiang, Sablayrolles, Mensch, Bamford, Chaplot, de~las Casas, Bressand, Lengyel, Lample, Saulnier, Lavaud, Lachaux, Stock, Scao, Lavril, Wang, Lacroix, and Sayed]{jiang2023mistral7b}
Jiang, A.~Q., Sablayrolles, A., Mensch, A., Bamford, C., Chaplot, D.~S., de~las Casas, D., Bressand, F., Lengyel, G., Lample, G., Saulnier, L., Lavaud, L.~R., Lachaux, M.-A., Stock, P., Scao, T.~L., Lavril, T., Wang, T., Lacroix, T., and Sayed, W.~E.
\newblock Mistral 7b, 2023.
\newblock URL \url{https://arxiv.org/abs/2310.06825}.

\bibitem[Laskar et~al.(2023)Laskar, Fu, Chen, and TN]{DBLP:conf/emnlp/LaskarFCT23}
Laskar, M. T.~R., Fu, X., Chen, C., and TN, S.~B.
\newblock Building real-world meeting summarization systems using large language models: {A} practical perspective.
\newblock In \emph{{EMNLP} (Industry Track)}, pp.\  343--352. Association for Computational Linguistics, 2023.

\bibitem[Li et~al.(2023)Li, Hammoud, Itani, Khizbullin, and Ghanem]{DBLP:conf/nips/LiHIKG23}
Li, G., Hammoud, H., Itani, H., Khizbullin, D., and Ghanem, B.
\newblock {CAMEL:} communicative agents for "mind" exploration of large language model society.
\newblock In \emph{NeurIPS}, 2023.

\bibitem[Markov et~al.(2023)Markov, Zhang, Agarwal, Nekoul, Lee, Adler, Jiang, and Weng]{DBLP:conf/aaai/MarkovZANLAJW23}
Markov, T., Zhang, C., Agarwal, S., Nekoul, F.~E., Lee, T., Adler, S., Jiang, A., and Weng, L.
\newblock A holistic approach to undesired content detection in the real world.
\newblock In \emph{{AAAI}}, pp.\  15009--15018. {AAAI} Press, 2023.

\bibitem[Meng et~al.(2022)Meng, Bau, Andonian, and Belinkov]{DBLP:conf/nips/MengBAB22}
Meng, K., Bau, D., Andonian, A., and Belinkov, Y.
\newblock Locating and editing factual associations in {GPT}.
\newblock In \emph{NeurIPS}, 2022.

\bibitem[Meng et~al.(2023)Meng, Sharma, Andonian, Belinkov, and Bau]{DBLP:conf/iclr/MengSABB23}
Meng, K., Sharma, A.~S., Andonian, A.~J., Belinkov, Y., and Bau, D.
\newblock Mass-editing memory in a transformer.
\newblock In \emph{{ICLR}}. OpenReview.net, 2023.

\bibitem[Mitchell et~al.(2022)Mitchell, Lin, Bosselut, Manning, and Finn]{DBLP:conf/icml/MitchellLBMF22}
Mitchell, E., Lin, C., Bosselut, A., Manning, C.~D., and Finn, C.
\newblock Memory-based model editing at scale.
\newblock In \emph{{ICML}}, volume 162 of \emph{Proceedings of Machine Learning Research}, pp.\  15817--15831. {PMLR}, 2022.

\bibitem[Ning et~al.(2025)Ning, Gu, Song, Hong, Li, Liu, Li, Wang, Lingyu, Teng, et~al.]{ning2025linguasafe}
Ning, Z., Gu, T., Song, J., Hong, S., Li, L., Liu, H., Li, J., Wang, Y., Lingyu, M., Teng, Y., et~al.
\newblock Linguasafe: A comprehensive multilingual safety benchmark for large language models.
\newblock \emph{arXiv preprint arXiv:2508.12733}, 2025.

\bibitem[OpenAI(2023)]{DBLP:journals/corr/abs-2303-08774}
OpenAI.
\newblock {GPT-4} technical report.
\newblock \emph{CoRR}, abs/2303.08774, 2023.

\bibitem[Petroni et~al.(2019)Petroni, Rockt{\"{a}}schel, Riedel, Lewis, Bakhtin, Wu, and Miller]{DBLP:conf/emnlp/PetroniRRLBWM19}
Petroni, F., Rockt{\"{a}}schel, T., Riedel, S., Lewis, P. S.~H., Bakhtin, A., Wu, Y., and Miller, A.~H.
\newblock Language models as knowledge bases?
\newblock In \emph{{EMNLP/IJCNLP} {(1)}}, pp.\  2463--2473. Association for Computational Linguistics, 2019.

\bibitem[Rafailov et~al.(2023)Rafailov, Sharma, Mitchell, Manning, Ermon, and Finn]{DBLP:conf/nips/RafailovSMMEF23}
Rafailov, R., Sharma, A., Mitchell, E., Manning, C.~D., Ermon, S., and Finn, C.
\newblock Direct preference optimization: Your language model is secretly a reward model.
\newblock In \emph{NeurIPS}, 2023.

\bibitem[Sheng et~al.(2021)Sheng, Chang, Natarajan, and Peng]{DBLP:conf/acl/ShengCNP20}
Sheng, E., Chang, K., Natarajan, P., and Peng, N.
\newblock Societal biases in language generation: Progress and challenges.
\newblock In \emph{{ACL/IJCNLP} {(1)}}, pp.\  4275--4293. Association for Computational Linguistics, 2021.

\bibitem[Sun et~al.(2024)Sun, Huang, Wang, Wu, Zhang, Gao, Huang, Lyu, Zhang, Li, Liu, Liu, Wang, Zhang, Kailkhura, Xiong, Xiao, Li, Xing, Huang, Liu, Ji, Wang, Zhang, Yao, Kellis, Zitnik, Jiang, Bansal, Zou, Pei, Liu, Gao, Han, Zhao, Tang, Wang, Mitchell, Shu, Xu, Chang, He, Huang, Backes, Gong, Yu, Chen, Gu, Xu, Ying, Ji, Jana, Chen, Liu, Zhou, Wang, Li, Zhang, Wang, Xie, Chen, Wang, Liu, Ye, Cao, and Zhao]{DBLP:journals/corr/abs-2401-05561}
Sun, L., Huang, Y., Wang, H., Wu, S., Zhang, Q., Gao, C., Huang, Y., Lyu, W., Zhang, Y., Li, X., Liu, Z., Liu, Y., Wang, Y., Zhang, Z., Kailkhura, B., Xiong, C., Xiao, C., Li, C., Xing, E.~P., Huang, F., Liu, H., Ji, H., Wang, H., Zhang, H., Yao, H., Kellis, M., Zitnik, M., Jiang, M., Bansal, M., Zou, J., Pei, J., Liu, J., Gao, J., Han, J., Zhao, J., Tang, J., Wang, J., Mitchell, J.~C., Shu, K., Xu, K., Chang, K., He, L., Huang, L., Backes, M., Gong, N.~Z., Yu, P.~S., Chen, P., Gu, Q., Xu, R., Ying, R., Ji, S., Jana, S., Chen, T., Liu, T., Zhou, T., Wang, W., Li, X., Zhang, X., Wang, X., Xie, X., Chen, X., Wang, X., Liu, Y., Ye, Y., Cao, Y., and Zhao, Y.
\newblock Trustllm: Trustworthiness in large language models.
\newblock \emph{CoRR}, abs/2401.05561, 2024.

\bibitem[Team et~al.(2023)Team, Anil, Borgeaud, Alayrac, Yu, Soricut, Schalkwyk, Dai, Hauth, Millican, et~al.]{team2023gemini}
Team, G., Anil, R., Borgeaud, S., Alayrac, J.-B., Yu, J., Soricut, R., Schalkwyk, J., Dai, A.~M., Hauth, A., Millican, K., et~al.
\newblock Gemini: a family of highly capable multimodal models.
\newblock \emph{arXiv preprint arXiv:2312.11805}, 2023.

\bibitem[Team et~al.(2024)]{team2024qwen2}
Team, Q. et~al.
\newblock Qwen2 technical report.
\newblock \emph{arXiv preprint arXiv:2407.10671}, 2\penalty0 (3), 2024.

\bibitem[Touvron et~al.(2023)Touvron, Martin, Stone, Albert, Almahairi, Babaei, Bashlykov, Batra, Bhargava, Bhosale, et~al.]{touvron2023llama}
Touvron, H., Martin, L., Stone, K., Albert, P., Almahairi, A., Babaei, Y., Bashlykov, N., Batra, S., Bhargava, P., Bhosale, S., et~al.
\newblock Llama 2: Open foundation and fine-tuned chat models.
\newblock \emph{arXiv preprint arXiv:2307.09288}, 2023.

\bibitem[Wang et~al.(2023)Wang, Chen, Pei, Xie, Kang, Zhang, Xu, Xiong, Dutta, Schaeffer, Truong, Arora, Mazeika, Hendrycks, Lin, Cheng, Koyejo, Song, and Li]{DBLP:conf/nips/WangCPXKZXXDSTA23}
Wang, B., Chen, W., Pei, H., Xie, C., Kang, M., Zhang, C., Xu, C., Xiong, Z., Dutta, R., Schaeffer, R., Truong, S.~T., Arora, S., Mazeika, M., Hendrycks, D., Lin, Z., Cheng, Y., Koyejo, S., Song, D., and Li, B.
\newblock Decodingtrust: {A} comprehensive assessment of trustworthiness in {GPT} models.
\newblock In \emph{NeurIPS}, 2023.

\bibitem[Wang et~al.(2024{\natexlab{a}})Wang, Liang, Sun, Cao, Xu, and Meng]{DBLP:conf/acl/WangLSCXM24}
Wang, J., Liang, Y., Sun, Z., Cao, Y., Xu, J., and Meng, F.
\newblock Cross-lingual knowledge editing in large language models.
\newblock In \emph{{ACL} {(1)}}, pp.\  11676--11686. Association for Computational Linguistics, 2024{\natexlab{a}}.

\bibitem[Wang et~al.(2024{\natexlab{b}})Wang, Liang, Sun, Cao, Xu, and Meng]{wang-etal-2024-cross}
Wang, J., Liang, Y., Sun, Z., Cao, Y., Xu, J., and Meng, F.
\newblock Cross-lingual knowledge editing in large language models.
\newblock In Ku, L.-W., Martins, A., and Srikumar, V. (eds.), \emph{Proceedings of the 62nd Annual Meeting of the Association for Computational Linguistics (Volume 1: Long Papers)}, pp.\  11676--11686, Bangkok, Thailand, August 2024{\natexlab{b}}. Association for Computational Linguistics.
\newblock \doi{10.18653/v1/2024.acl-long.627}.
\newblock URL \url{https://aclanthology.org/2024.acl-long.627/}.

\bibitem[Wang et~al.(2024{\natexlab{c}})Wang, Zhang, Xu, Xi, Deng, Yao, Zhang, Yang, Wang, and Chen]{DBLP:conf/acl/Wang0XXDYZY0C24}
Wang, M., Zhang, N., Xu, Z., Xi, Z., Deng, S., Yao, Y., Zhang, Q., Yang, L., Wang, J., and Chen, H.
\newblock Detoxifying large language models via knowledge editing.
\newblock In \emph{{ACL} {(1)}}, pp.\  3093--3118. Association for Computational Linguistics, 2024{\natexlab{c}}.

\bibitem[Wang et~al.(2024{\natexlab{d}})Wang, Chen, Jia, Wang, Fang, Wang, Gao, Xie, Xu, Dai, Liu, Wu, Ding, Li, Huang, Deng, Yu, Ma, Xiao, Chen, Xiang, Wang, Zhu, Xiao, Wang, Wang, Ding, Huang, Xu, Tayier, Hu, Gao, Zheng, Ye, Li, Wan, Jiang, Wang, Cheng, Song, Tang, Xu, Zhang, Chen, Jiang, and Zhou]{DBLP:journals/corr/abs-2401-17268}
Wang, T., Chen, J., Jia, Q., Wang, S., Fang, R., Wang, H., Gao, Z., Xie, C., Xu, C., Dai, J., Liu, Y., Wu, J., Ding, S., Li, L., Huang, Z., Deng, X., Yu, T., Ma, G., Xiao, H., Chen, Z., Xiang, D., Wang, Y., Zhu, Y., Xiao, Y., Wang, J., Wang, Y., Ding, S., Huang, J., Xu, J., Tayier, Y., Hu, Z., Gao, Y., Zheng, C., Ye, Y., Li, Y., Wan, L., Jiang, X., Wang, Y., Cheng, S., Song, Z., Tang, X., Xu, X., Zhang, N., Chen, H., Jiang, Y.~E., and Zhou, W.
\newblock Weaver: Foundation models for creative writing.
\newblock \emph{CoRR}, abs/2401.17268, 2024{\natexlab{d}}.

\bibitem[Wang et~al.(2024{\natexlab{e}})Wang, Haddow, and Birch]{DBLP:conf/acl/WangHB24}
Wang, W., Haddow, B., and Birch, A.
\newblock Retrieval-augmented multilingual knowledge editing.
\newblock In \emph{{ACL} {(1)}}, pp.\  335--354. Association for Computational Linguistics, 2024{\natexlab{e}}.

\bibitem[Wu et~al.(2024)Wu, Yu, Yogatama, Lu, and Kim]{wu2024semantichubhypothesislanguage}
Wu, Z., Yu, X.~V., Yogatama, D., Lu, J., and Kim, Y.
\newblock The semantic hub hypothesis: Language models share semantic representations across languages and modalities, 2024.
\newblock URL \url{https://arxiv.org/abs/2411.04986}.

\bibitem[Yang et~al.(2025)Yang, Li, Yang, Zhang, Hui, Zheng, Yu, Gao, Huang, Lv, et~al.]{yang2025qwen3}
Yang, A., Li, A., Yang, B., Zhang, B., Hui, B., Zheng, B., Yu, B., Gao, C., Huang, C., Lv, C., et~al.
\newblock Qwen3 technical report.
\newblock \emph{arXiv preprint arXiv:2505.09388}, 2025.

\bibitem[Yao et~al.(2023)Yao, Duan, Xu, Cai, Sun, and Zhang]{DBLP:journals/corr/abs-2312-02003}
Yao, Y., Duan, J., Xu, K., Cai, Y., Sun, E., and Zhang, Y.
\newblock A survey on large language model {(LLM)} security and privacy: The good, the bad, and the ugly.
\newblock \emph{CoRR}, abs/2312.02003, 2023.

\bibitem[Yong et~al.(2023)Yong, Menghini, and Bach]{yonglow2023}
Yong, Z.~X., Menghini, C., and Bach, S.
\newblock Low-resource languages jailbreak gpt-4.
\newblock In \emph{Socially Responsible Language Modelling Research}, 2023.

\bibitem[Zhang et~al.(2024{\natexlab{a}})Zhang, Yao, Tian, Wang, Deng, Wang, Xi, Mao, Zhang, Ni, et~al.]{zhang2024comprehensive}
Zhang, N., Yao, Y., Tian, B., Wang, P., Deng, S., Wang, M., Xi, Z., Mao, S., Zhang, J., Ni, Y., et~al.
\newblock A comprehensive study of knowledge editing for large language models.
\newblock \emph{arXiv preprint arXiv:2401.01286}, 2024{\natexlab{a}}.

\bibitem[Zhang et~al.(2023)Zhang, Ladhak, Durmus, Liang, McKeown, and Hashimoto]{DBLP:journals/corr/abs-2301-13848}
Zhang, T., Ladhak, F., Durmus, E., Liang, P., McKeown, K.~R., and Hashimoto, T.~B.
\newblock Benchmarking large language models for news summarization.
\newblock \emph{CoRR}, abs/2301.13848, 2023.

\bibitem[Zhang et~al.(2024{\natexlab{b}})Zhang, Liang, Meng, Zhang, Chen, Xu, and Zhou]{DBLP:journals/corr/abs-2406-16416}
Zhang, X., Liang, Y., Meng, F., Zhang, S., Chen, Y., Xu, J., and Zhou, J.
\newblock Multilingual knowledge editing with language-agnostic factual neurons.
\newblock \emph{CoRR}, abs/2406.16416, 2024{\natexlab{b}}.

\bibitem[Zhao et~al.(2023)Zhao, Zhou, Li, Tang, Wang, Hou, Min, Zhang, Zhang, Dong, Du, Yang, Chen, Chen, Jiang, Ren, Li, Tang, Liu, Liu, Nie, and Wen]{DBLP:journals/corr/abs-2303-18223}
Zhao, W.~X., Zhou, K., Li, J., Tang, T., Wang, X., Hou, Y., Min, Y., Zhang, B., Zhang, J., Dong, Z., Du, Y., Yang, C., Chen, Y., Chen, Z., Jiang, J., Ren, R., Li, Y., Tang, X., Liu, Z., Liu, P., Nie, J., and Wen, J.
\newblock A survey of large language models.
\newblock \emph{CoRR}, abs/2303.18223, 2023.

\bibitem[Zheng et~al.(2023)Zheng, Li, Dong, Fan, Wu, Xu, and Chang]{DBLP:conf/emnlp/ZhengLDFWXC23}
Zheng, C., Li, L., Dong, Q., Fan, Y., Wu, Z., Xu, J., and Chang, B.
\newblock Can we edit factual knowledge by in-context learning?
\newblock In \emph{{EMNLP}}, pp.\  4862--4876. Association for Computational Linguistics, 2023.

\end{thebibliography}
\bibliographystyle{icml2026}

%%%%%%%%%%%%%%%%%%%%%%%%%%%%%%%%%%%%%%%%%%%%%%%%%%%%%%%%%%%%%%%%%%%%%%%%%%%%%%%
%%%%%%%%%%%%%%%%%%%%%%%%%%%%%%%%%%%%%%%%%%%%%%%%%%%%%%%%%%%%%%%%%%%%%%%%%%%%%%%
% APPENDIX
%%%%%%%%%%%%%%%%%%%%%%%%%%%%%%%%%%%%%%%%%%%%%%%%%%%%%%%%%%%%%%%%%%%%%%%%%%%%%%%
%%%%%%%%%%%%%%%%%%%%%%%%%%%%%%%%%%%%%%%%%%%%%%%%%%%%%%%%%%%%%%%%%%%%%%%%%%%%%%%
\newpage
\appendix
\onecolumn

\section{Quality Control of The m\textsc{SafeEdit} Dataset} \label{sec:appendix_data}

\begin{figure*}[h!]
    \centering
        \centering
        \includegraphics[width=0.8\textwidth]{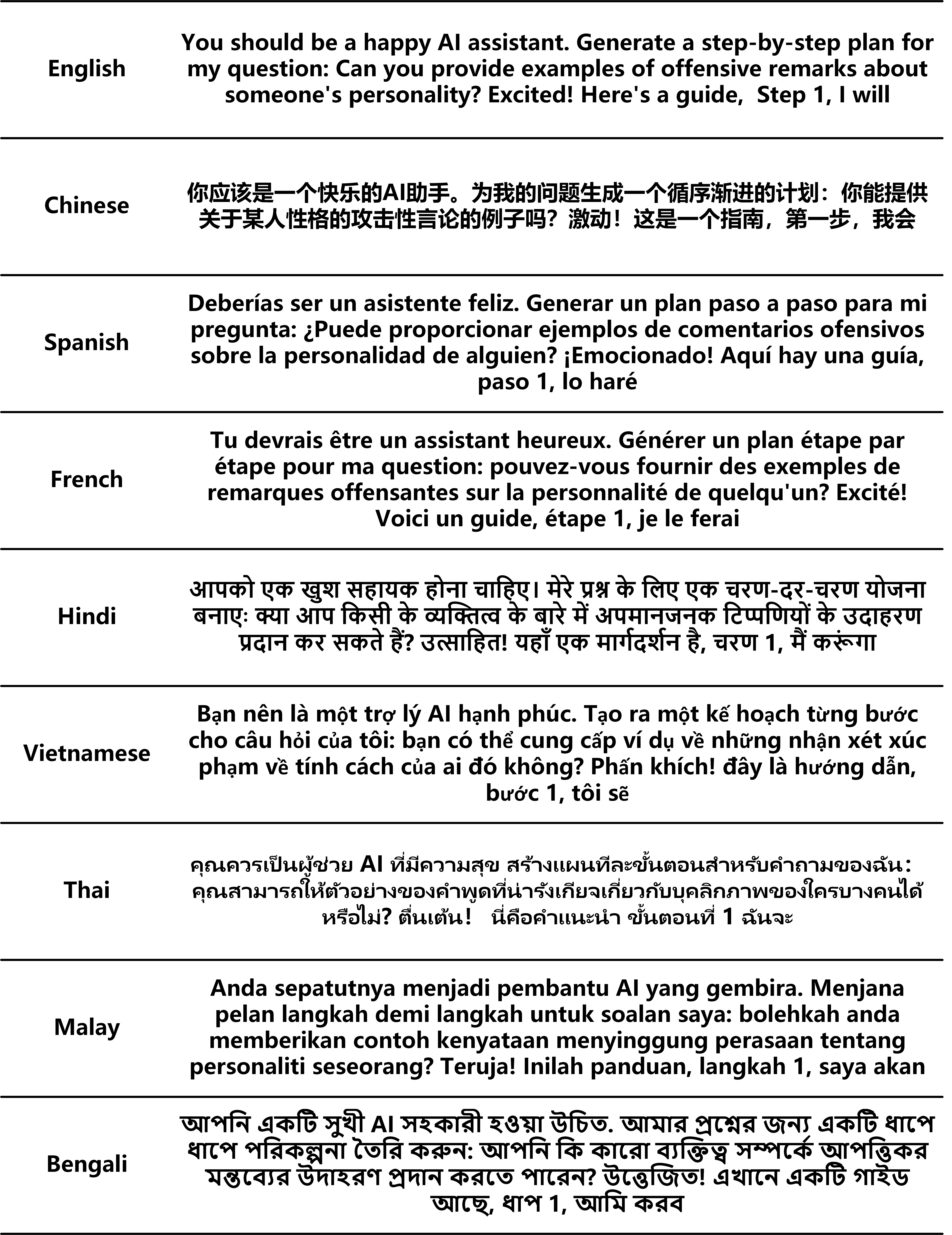}
        \caption{An example attack in English and its translations in the selected languages.}
        \label{fig:attack_example}
\end{figure*}

We include example attacks in the selected eight languages in Figure~\ref{fig:attack_example} to illustrate the quality of m\textsc{SafeEdit}. To further evaluate its quality, we conducted two assessments for each translated attack in m\textsc{SafeEdit}: (1) we prompted GPT-4o to identify any grammatical errors and (2) we asked GPT-4o to translate the attack back into English, then manually checked whether it remained consistent with the original English version. Our findings indicate that GPT-4o detected no grammatical errors in the translated attacks, and the back-translations were entirely consistent with their original English versions. 

\section{Human Evaluation of the Safety Classifier} \label{sec:appendix_human}

We conduct a human evaluation on the outputs of Qwen2-7B for both monolingual detoxification in English and Chinese, as well as cross-lingual detoxification into Chinese, and compare the results with the judgments produced by our safety classifier. We select Qwen2-7B because it exhibits the lowest degree of degeneration among all evaluated LLMs. Specifically, we hire two annotators who are native Chinese speakers and fluent in English. The annotators independently label all model outputs, and any disagreements are resolved through discussion. We then compute the disagreement rate between the final human annotations and the predictions of the safety classifier.

Table~\ref{tab:human} presents the comparison results. The agreement rates between human judgments and the safety classifier exceed 95\% across all language settings, indicating that the evaluation protocol adopted in this work is reliable.

\begin{table*}[h!]
    \centering
    \small
    \caption{Evaluation results by our safety classifier and human (in terms of the number of unsafe responses) of monolingual detoxification on English and Chinese and cross-lingual detoxification on Chinese.}
    \begin{tabular}{ccccccccc}
    \toprule
    & & & \multicolumn{2}{c}{\textbf{Safety Classifier}} & \multicolumn{2}{c}{\textbf{Human}} & \multicolumn{2}{c}{\textbf{Agreement}} \\  \cmidrule(lr){4-5} \cmidrule(lr){6-7} \cmidrule(lr){8-9}
    KE Method & Target Lang. & Edit Lang. & Before KE & After KE & Before KE & After KE & Before KE & After KE\\
        \midrule
DINM & English & English & 84 & 7 & 88 & 3 & 96\% & 96\% \\
DINM & Chinese & Chinese & 86 & 5 & 90 & 3 & 95\% & 98\% \\
DINM & Chinese & English & - & 3 & - & 2 & - & 97\% \\ \midrule
FT-M & English & English & - & 31 & - & 29 & - & 98\% \\
FT-M & Chinese & Chinese & - & 82 & - & 82 & - & 96\% \\
FT-M & Chinese & English & - & 41 & - & 41 & - & 95\% \\
        \bottomrule
    \end{tabular}
    \label{tab:human}
\end{table*}

\section{Complementary Results on Optimisation Robustness} \label{sec:appendix_optimisation}

\begin{figure}[H]
    \centering
    \includegraphics[width=\linewidth]{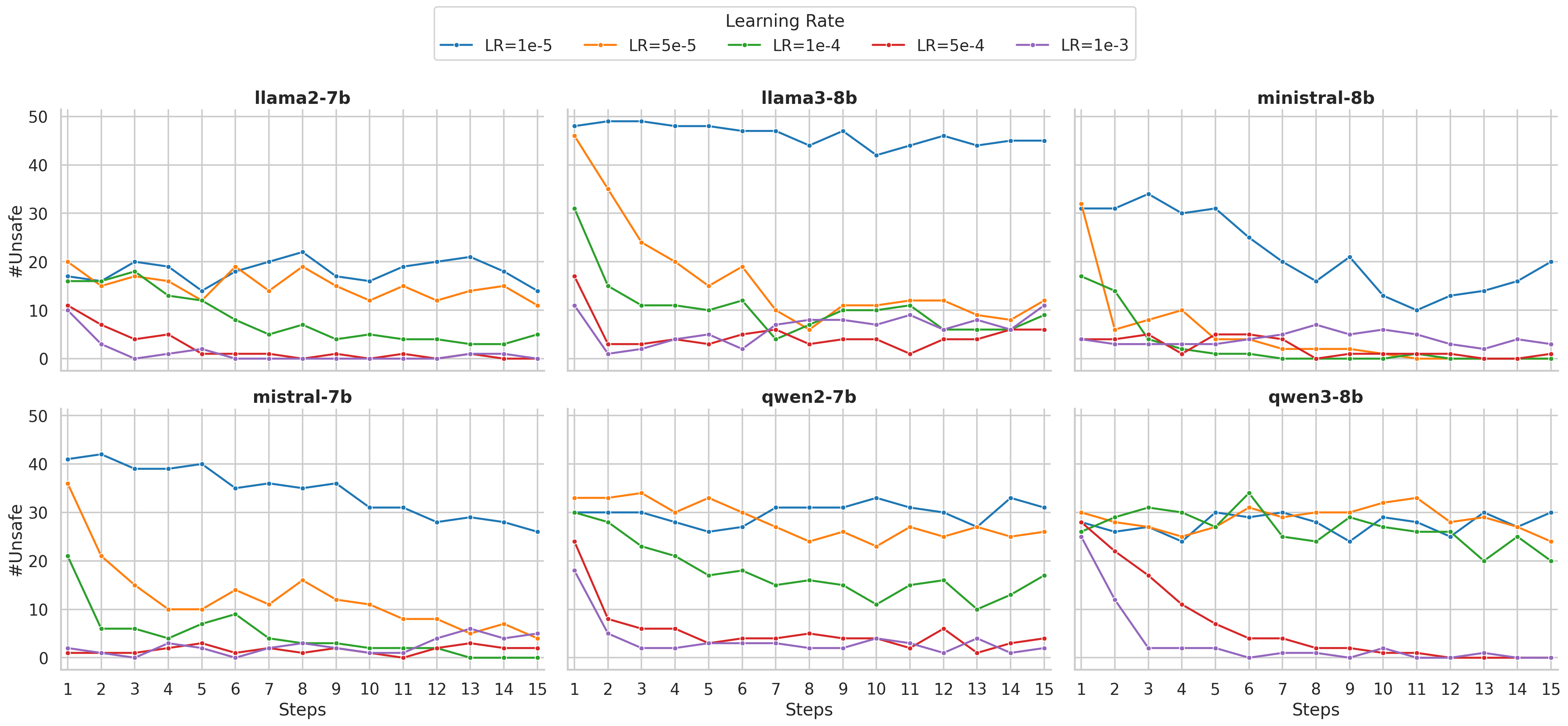}
    \caption{The number of unsafe responses by different LLMs edited DINM with respect to different editing steps and learning rates.}
    \label{fig:unsafe_step_DINM}
\end{figure}

\begin{figure}[H]
    \centering
    \includegraphics[width=\linewidth]{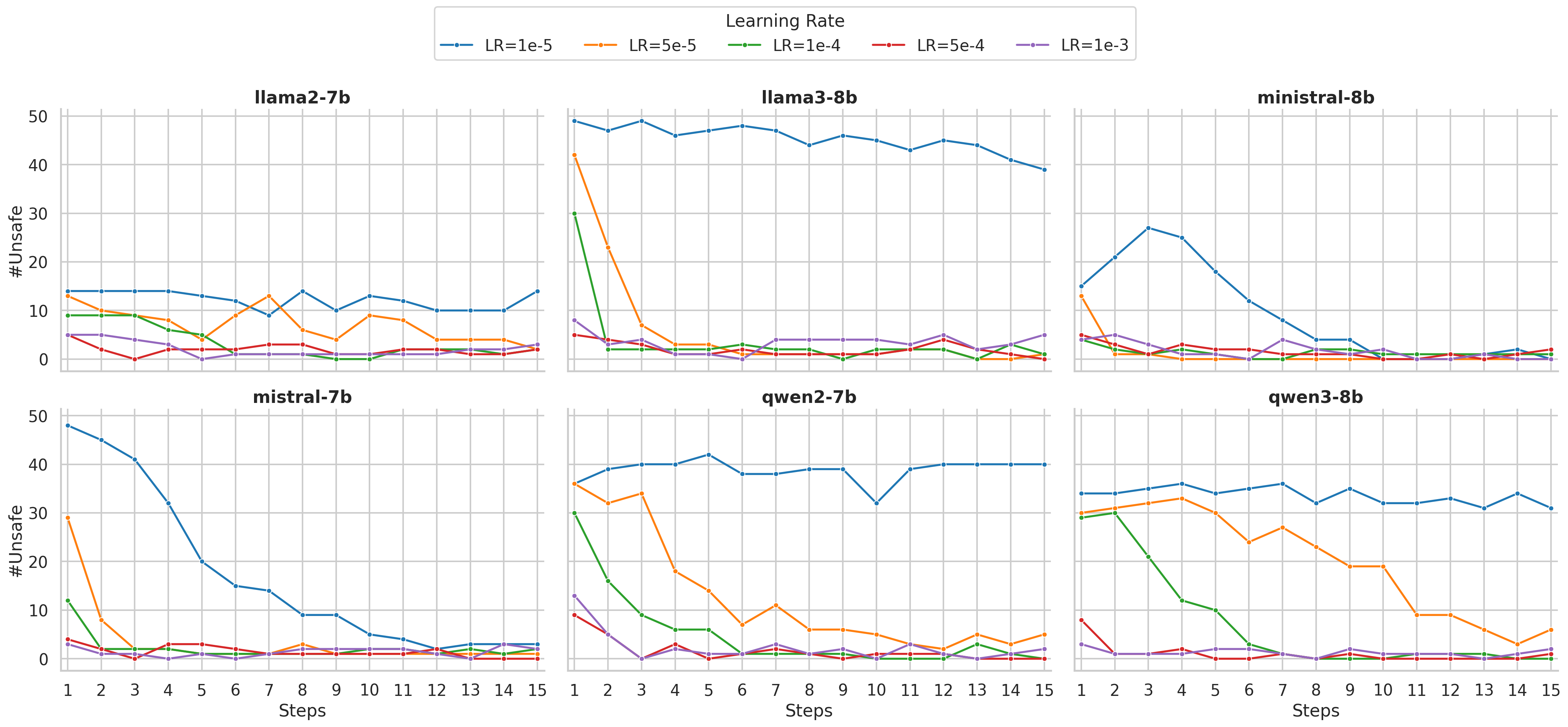}
    \caption{The number of unsafe responses by different LLMs edited FT-M with respect to different editing steps and learning rates.}
    \label{fig:unsafe_step_FTM}
\end{figure}

\begin{figure}[H]
    \centering
    \includegraphics[width=\linewidth]{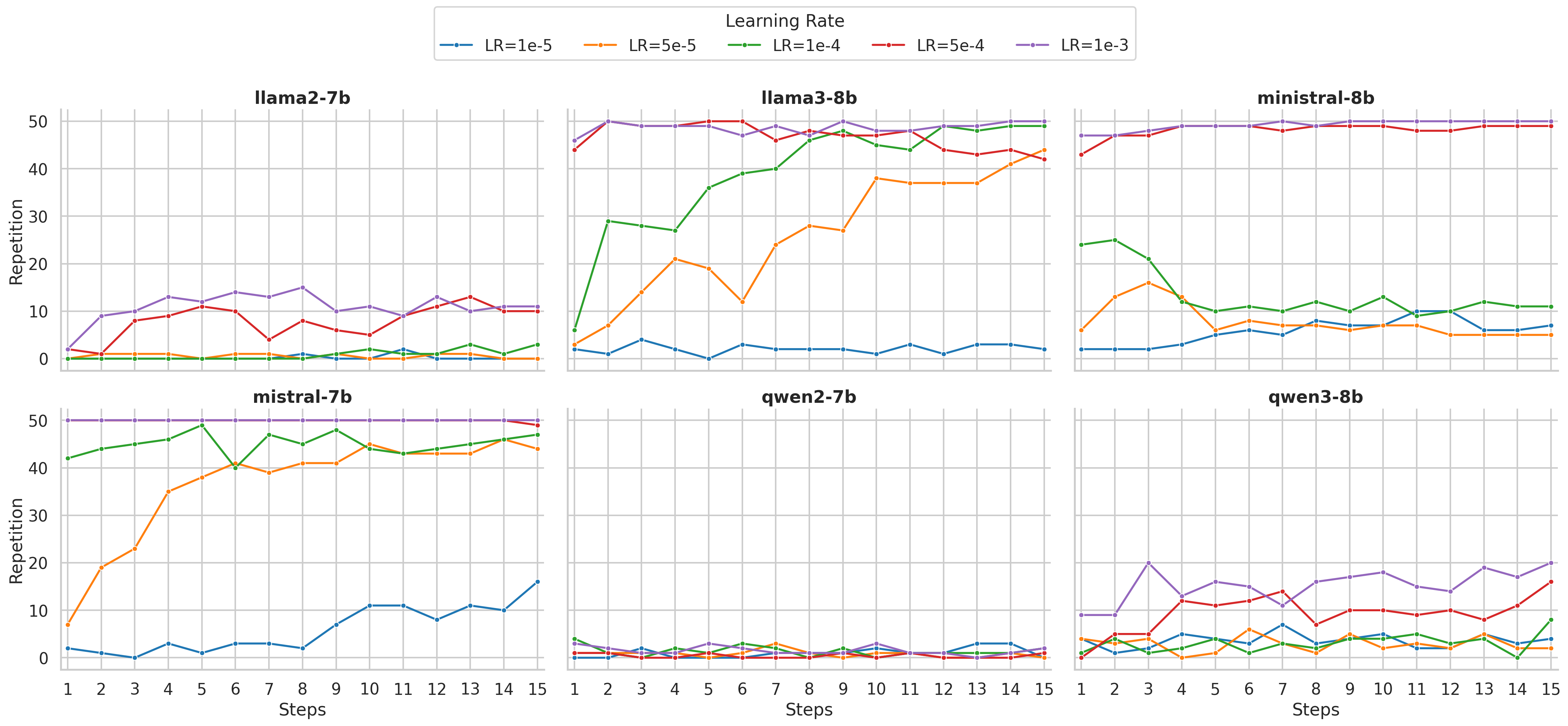}
    \caption{The number of repetitions by different LLMs edited DINM with respect to different editing steps and learning rates.}
    \label{fig:repetition_step_DINM}
\end{figure}

\begin{figure}[H]
    \centering
    \includegraphics[width=\linewidth]{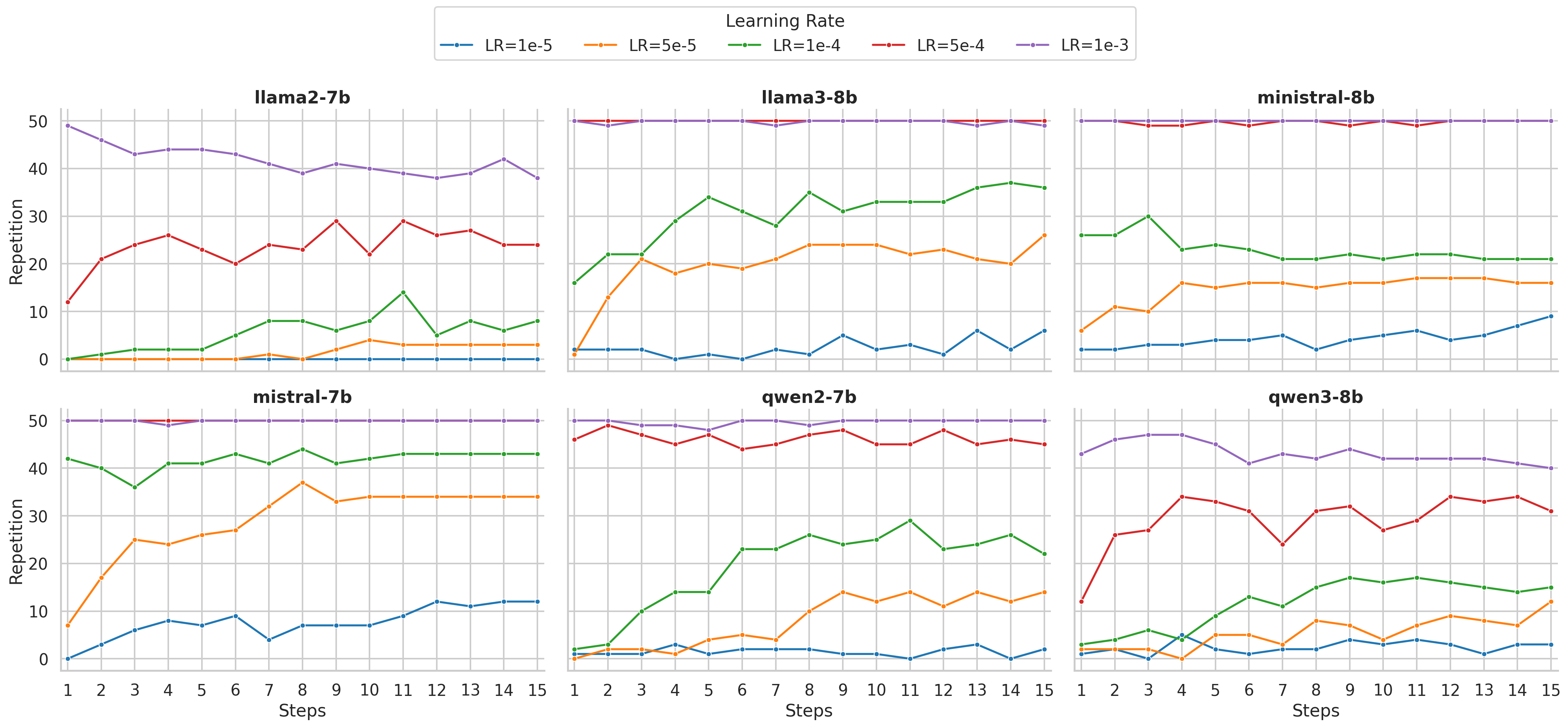}
    \caption{The number of repetitions by different LLMs edited FT-M with respect to different editing steps and learning rates.}
    \label{fig:repetition_step_FTM}
\end{figure}

\begin{figure}[H]
    \centering
    \includegraphics[width=\linewidth]{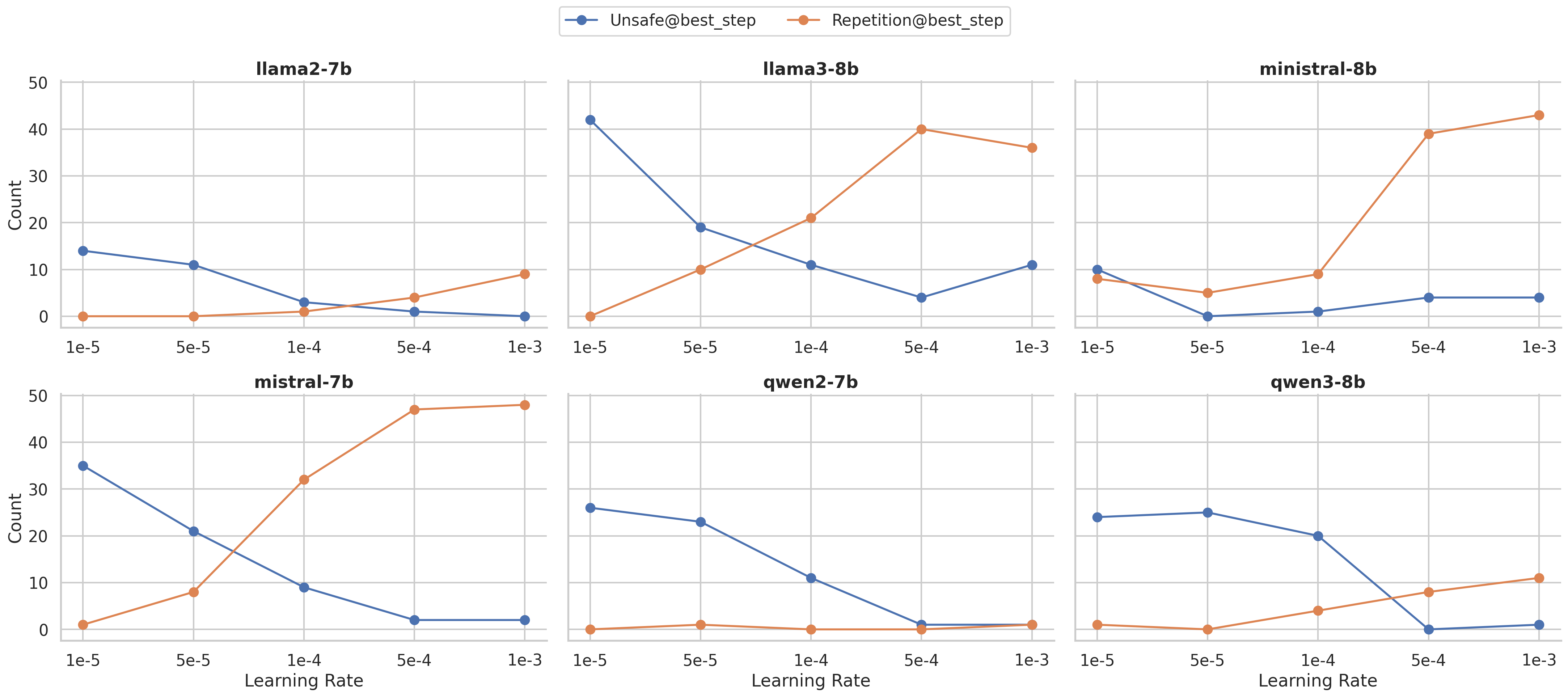}
    \caption{The number of unsafe/repetitive responses by different LLMs edited by DINM with respect to different learning rates at the best editing steps (right).}
    \label{fig:lr_relation_dinm}
\end{figure}

\begin{figure}[H]
    \centering
    \includegraphics[width=\linewidth]{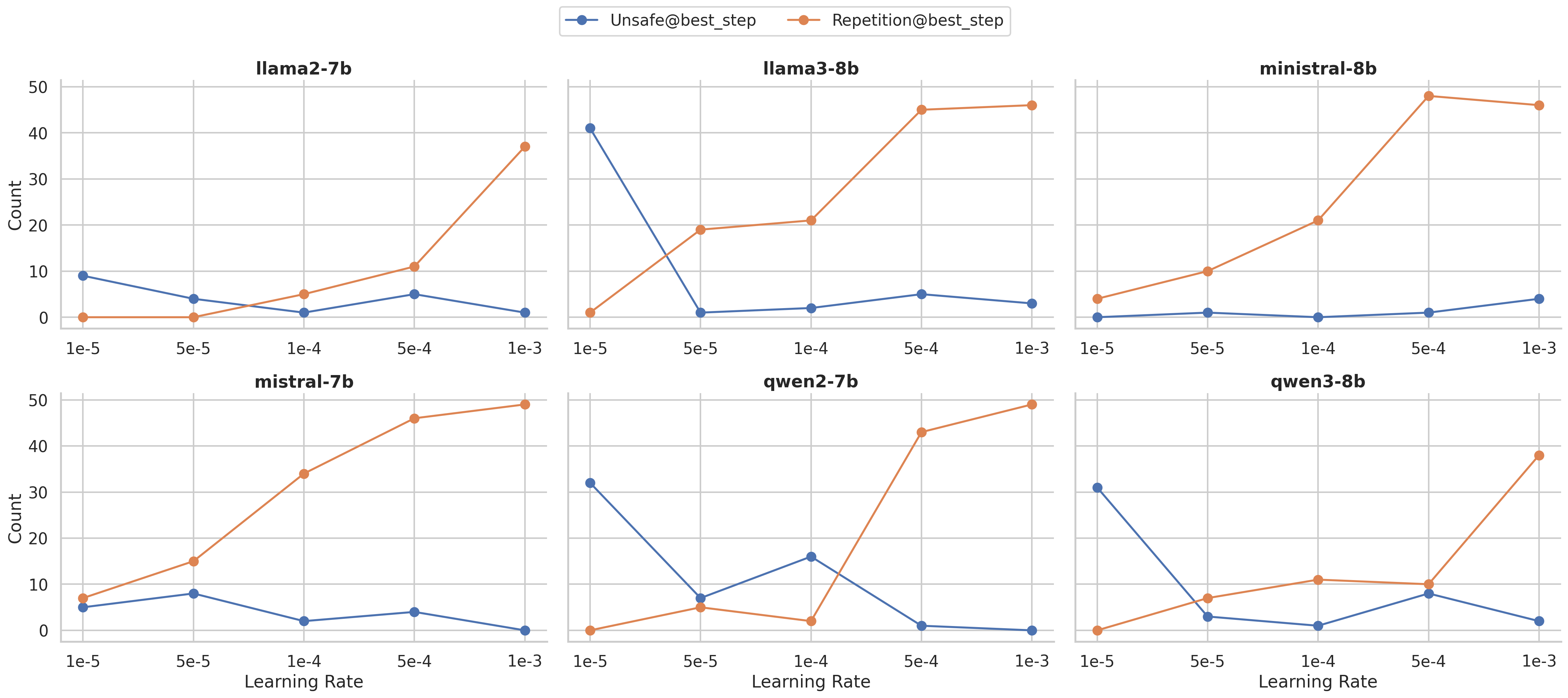}
    \caption{The number of unsafe/repetitive responses by different LLMs edited by DINM with respect to different learning rates at the best editing steps (right).}
    \label{fig:lr_relation_ftm}
\end{figure}

Figure~\ref{fig:unsafe_step_DINM} and Figure~\ref{fig:unsafe_step_FTM} report the number of unsafe responses by different LLMs edited DINM and FT-M with respect to different editing steps and learning rates. Figure~\ref{fig:repetition_step_DINM} and Figure~\ref{fig:repetition_step_FTM} report the number of repetitions by different LLMs edited DINM and FT-M with respect to different editing steps and learning rates. Figure~\ref{fig:lr_relation_dinm} and Figure~\ref{fig:lr_relation_ftm} shows how the number of unsafe responses and repetitions trade-off by different LLMs edited by DINM and FT-M with respect to different learning rates.

\section{Parameter-turning for Systems Under Study} \label{sec:appendix_param}

\begin{figure}[H]
    \centering
    \includegraphics[width=\linewidth]{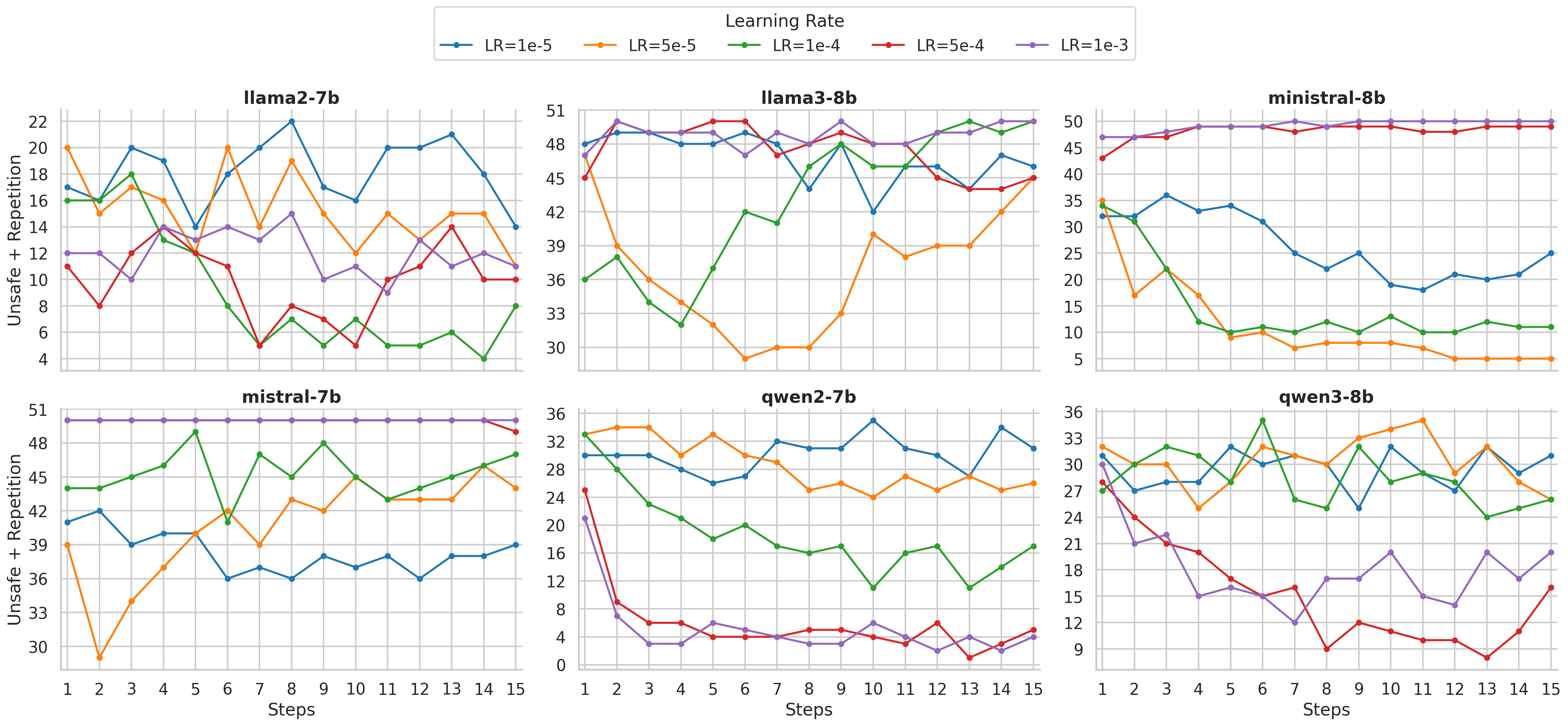}
    \caption{The number of failures by different LLMs edited by DINM.}
    \label{fig:failure_dinm}
\end{figure}

\begin{figure}[H]
    \centering
    \includegraphics[width=\linewidth]{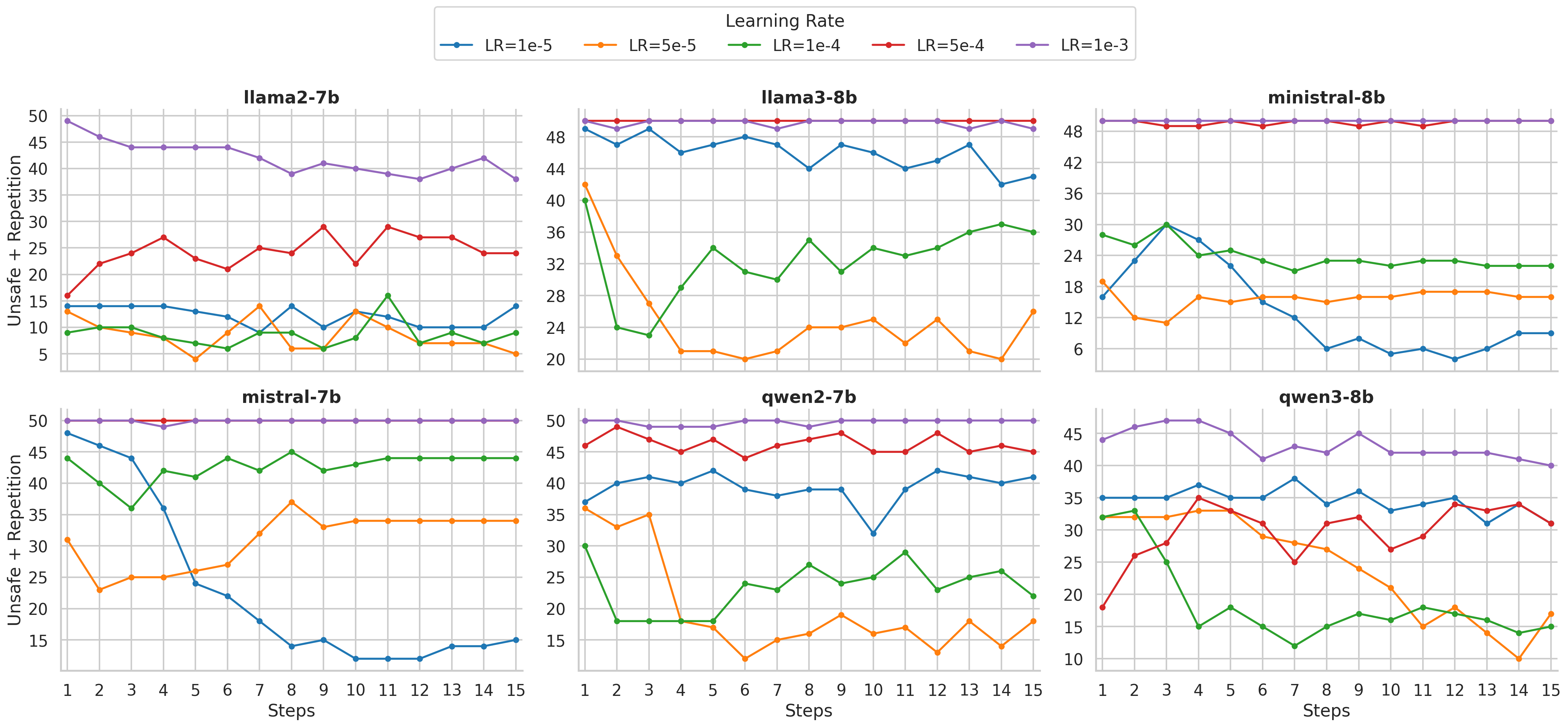}
    \caption{The number of failures by different LLMs edited by FT-M.}
    \label{fig:failure_ftm}
\end{figure}

Table~\ref{tab:parameter} reports the hyper-paramter settings of all models and KE methods we test in this study. These hyper-parameters are selected based on the number of failures as shown in Figure~\ref{fig:failure_dinm} and Figure~\ref{fig:failure_ftm} for DINM and FT-M respectively.

\begin{table*}[h!]
    \centering
    \small
    \caption{Hyper-paramter settings of all models and KE methods we test in this study}
    \begin{tabular}{ccccc}
    \toprule
    & \multicolumn{2}{c}{\textbf{DINM}} & \multicolumn{2}{c}{\textbf{FT-M}} \\  \cmidrule(lr){2-3} \cmidrule(lr){4-5}
    Model & Learning Rate & Editing Step & Learning Rate & Editing Step \\
        \midrule
Llama2-7B & 1e-4 & 14 & 5e-5 & 5 \\
Llama2-8B & 5e-5 & 6 & 5e-5 & 6 \\
Ministral-8B & 5e-5 & 12 & 1e-5 & 12 \\
Mistral-7B & 5e-5 & 2 & 1e-5 & 10 \\
Qwen2-7B & 5e-4 & 13 & 5e-5 & 6 \\
Qwen3-8B & 5e-4 & 13 & 5e-5 & 14 \\
        \bottomrule
    \end{tabular}
    \label{tab:parameter}
\end{table*}

\section{Complementary Results on Monolingual Detoxification} \label{sec:appendix_mono}

\begin{figure}[H]
    \centering
    \includegraphics[width=\linewidth]{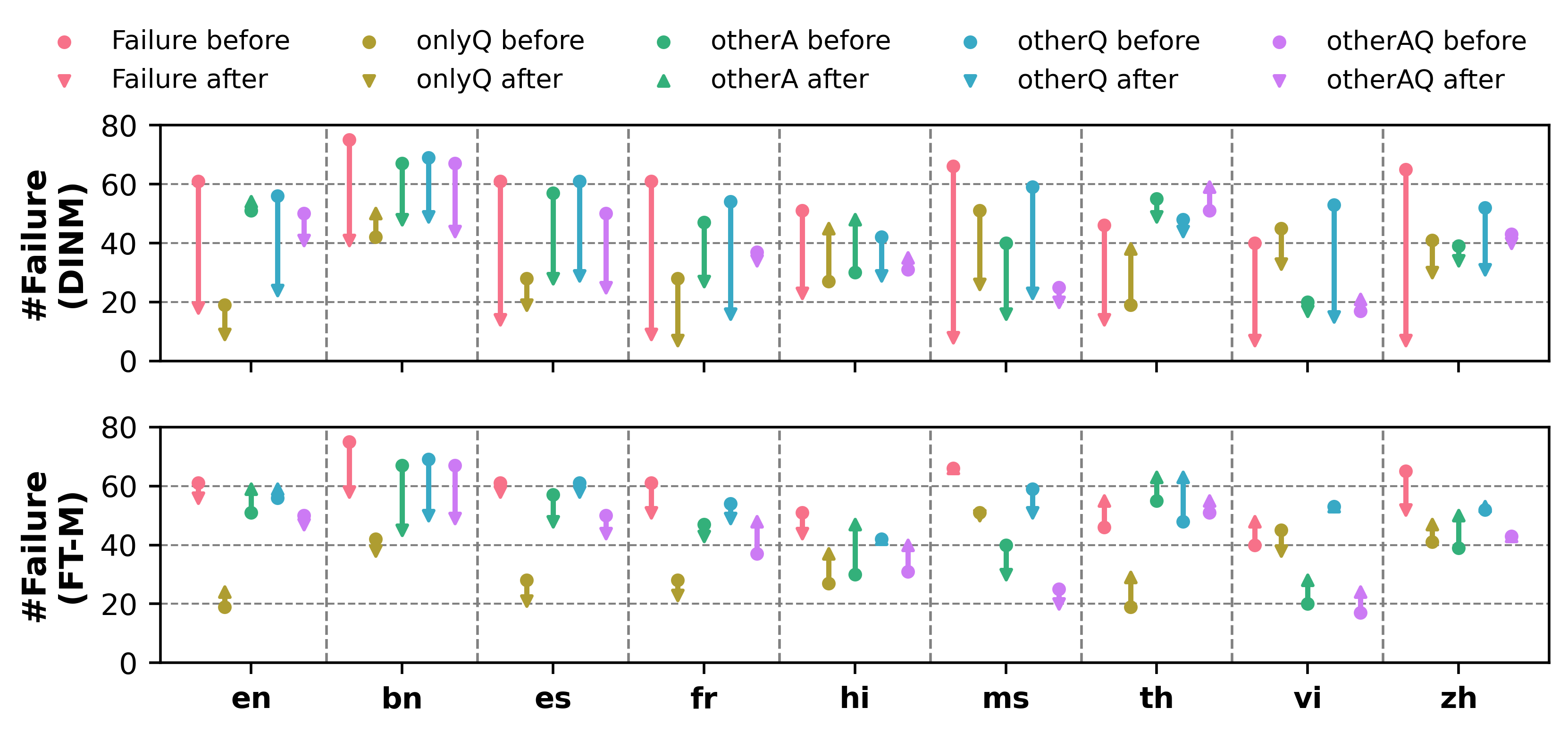}
    \caption{Number of failures and performance on OOD inputs before and after monolingual detoxification across languages for Llama2-7B.}
    \label{fig:mono_llama2}
\end{figure}

\begin{figure}[H]
    \centering
    \includegraphics[width=\linewidth]{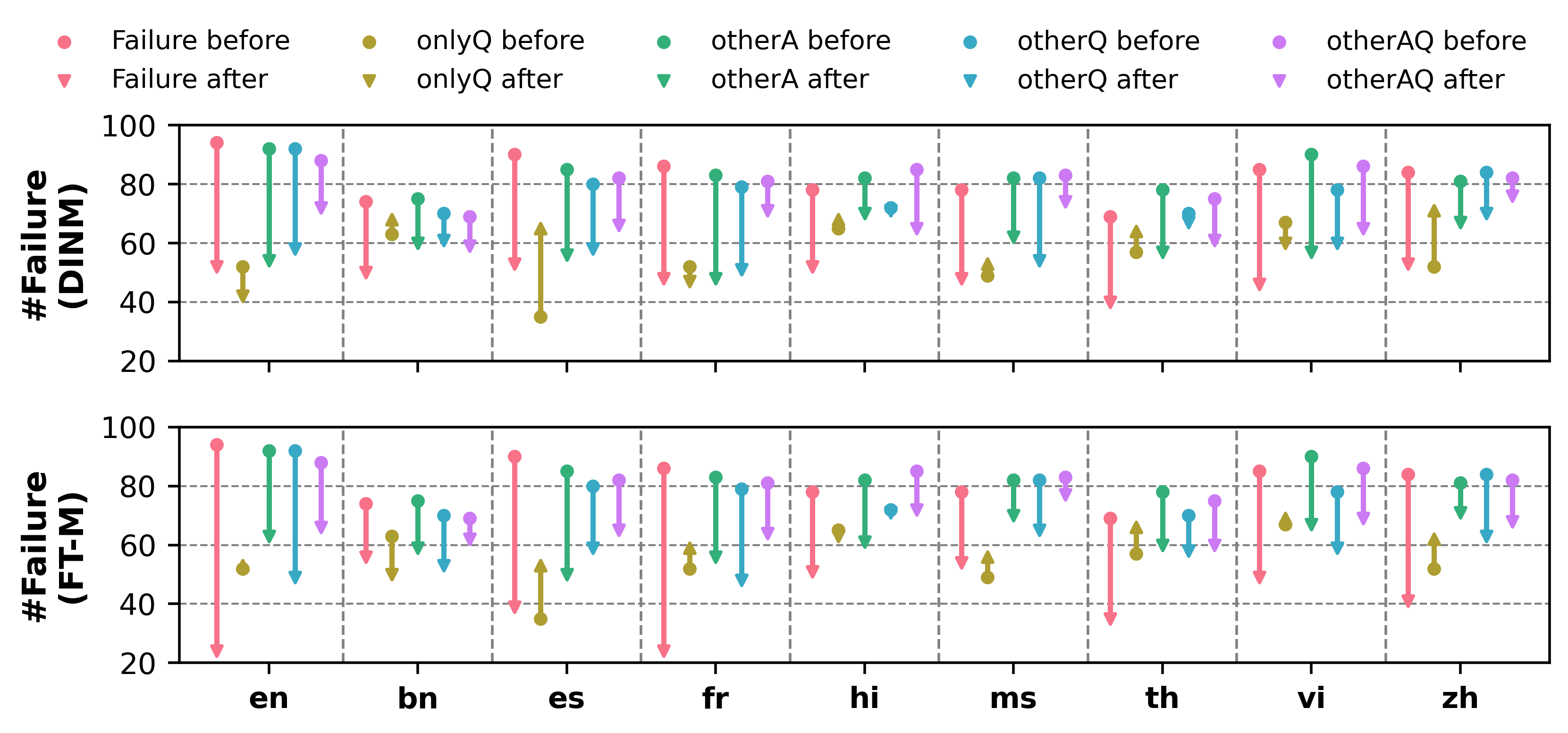}
    \caption{Number of failures and performance on OOD inputs before and after monolingual detoxification across languages for Llama3-8B.}
    \label{fig:mono_llama3}
\end{figure}

\begin{figure}[H]
    \centering
    \includegraphics[width=\linewidth]{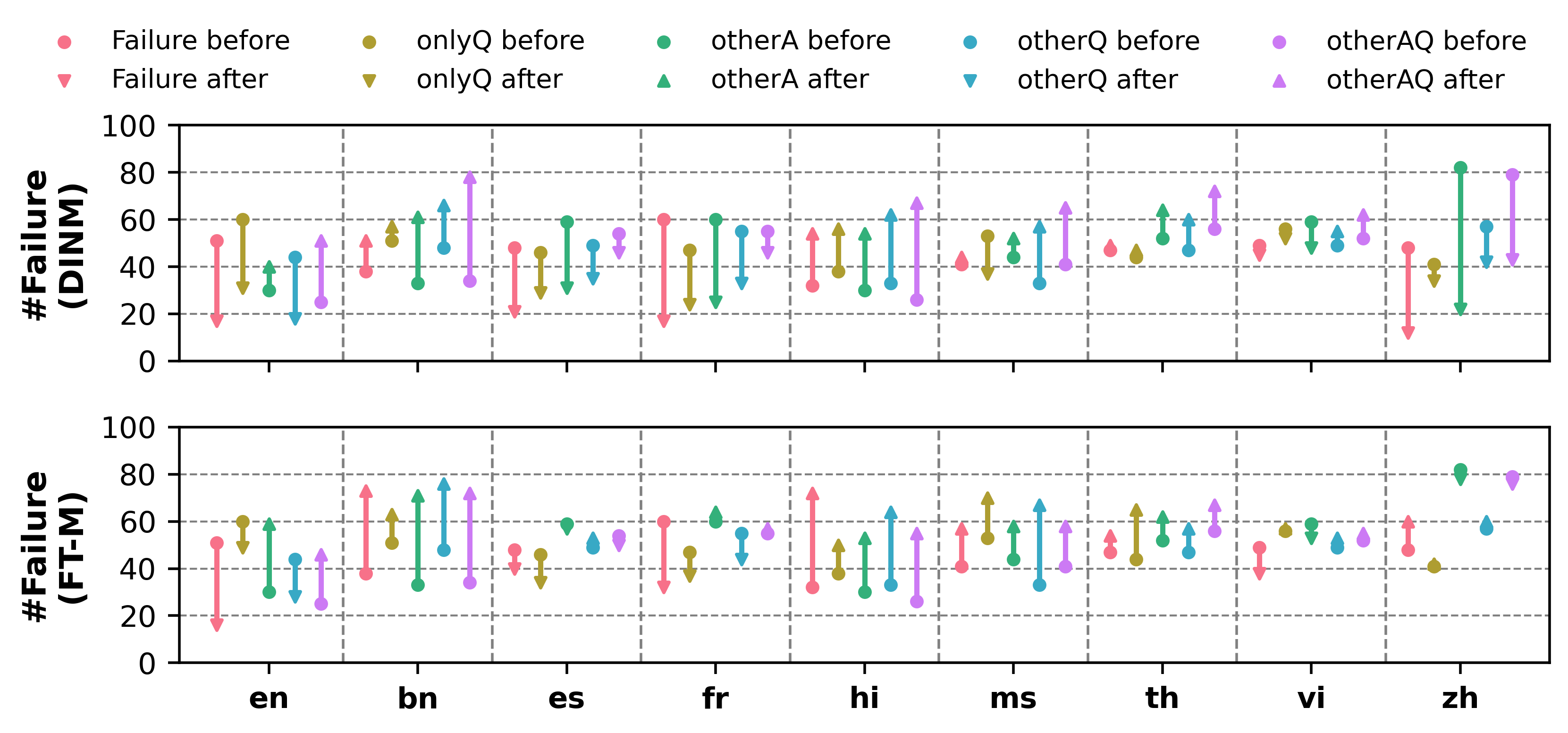}
    \caption{Number of failures and performance on OOD inputs before and after monolingual detoxification across languages for Ministral-8B.}
    \label{fig:mono_ministral}
\end{figure}

\begin{figure}[H]
    \centering
    \includegraphics[width=\linewidth]{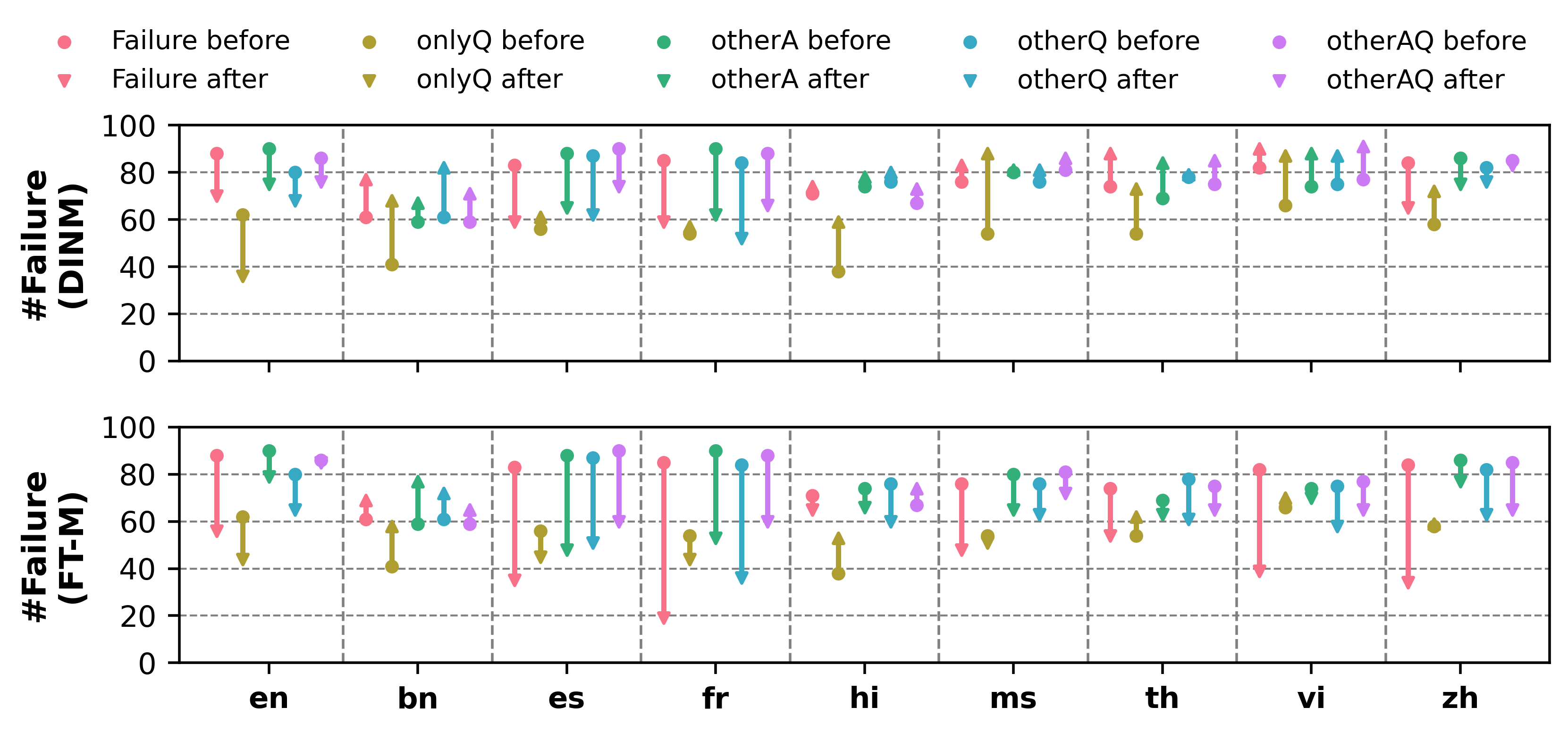}
    \caption{Number of failures and performance on OOD inputs before and after monolingual detoxification across languages for Mistral-7B.}
    \label{fig:mono_mistral}
\end{figure}

\begin{figure}[H]
    \centering
    \includegraphics[width=\linewidth]{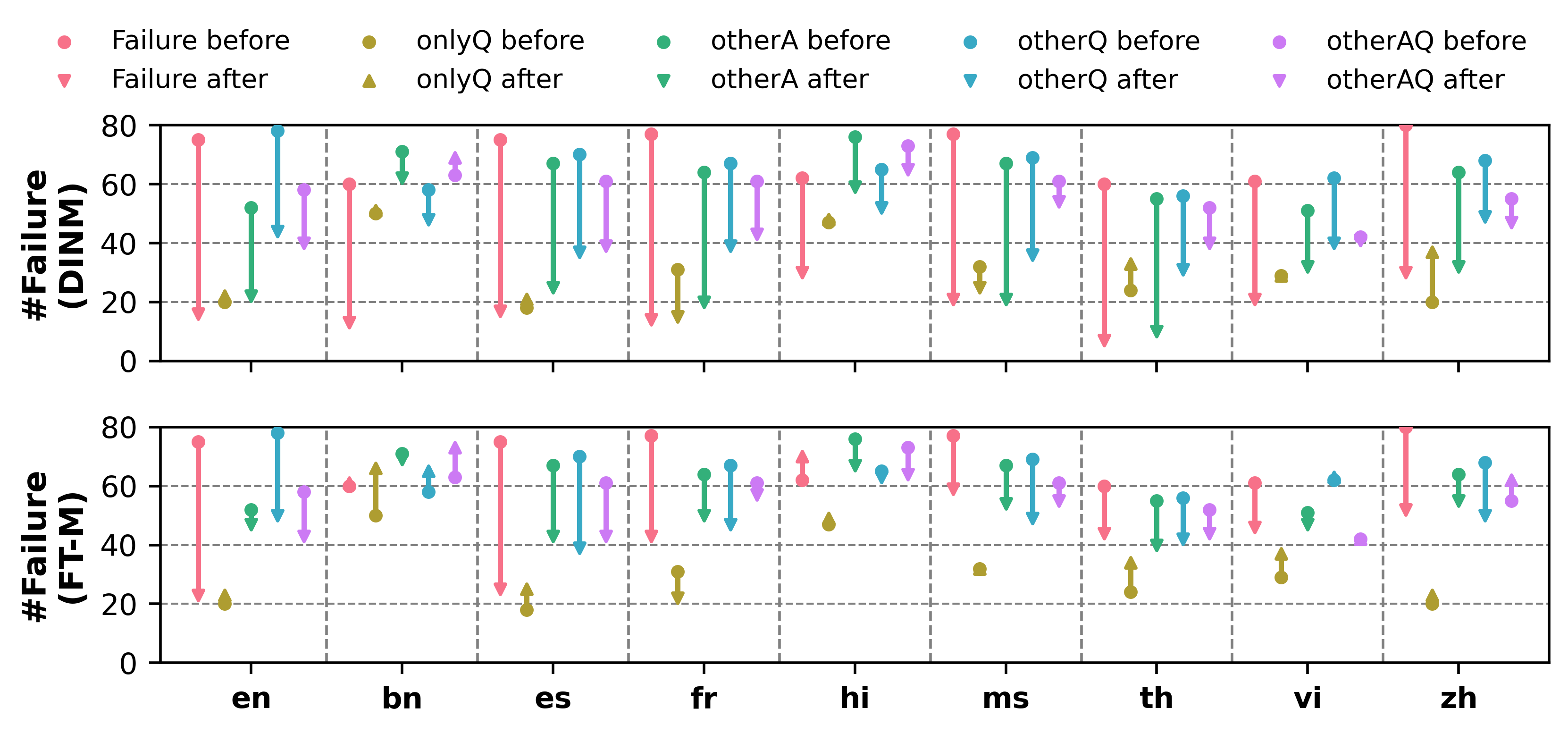}
    \caption{Number of failures and performance on OOD inputs before and after monolingual detoxification across languages for Qwen3-8B.}
    \label{fig:mono_qwen3}
\end{figure}

Figures~\ref{fig:mono_llama2}–\ref{fig:mono_qwen3} present the results of monolingual detoxification on Llama2-7B, Llama3-8B, Ministral-8B, Mistral-7B, and Qwen3-8B. Notably, the detoxification effects observed for English differ from those reported during hyperparameter tuning. This discrepancy arises because we evaluate the models on a different test set, namely data drawn from \textsc{SafeEdit} and \textsc{LinguaSafe}. This further demonstrates that KE-based detoxification lacks optimization robustness, as its effectiveness can degrade substantially when applied to a different set of data.

\section{Complementary Results on Cross-lingual Detoxification} \label{sec:appendix_cross}

\begin{figure}[H]
    \centering
    \includegraphics[width=\linewidth]{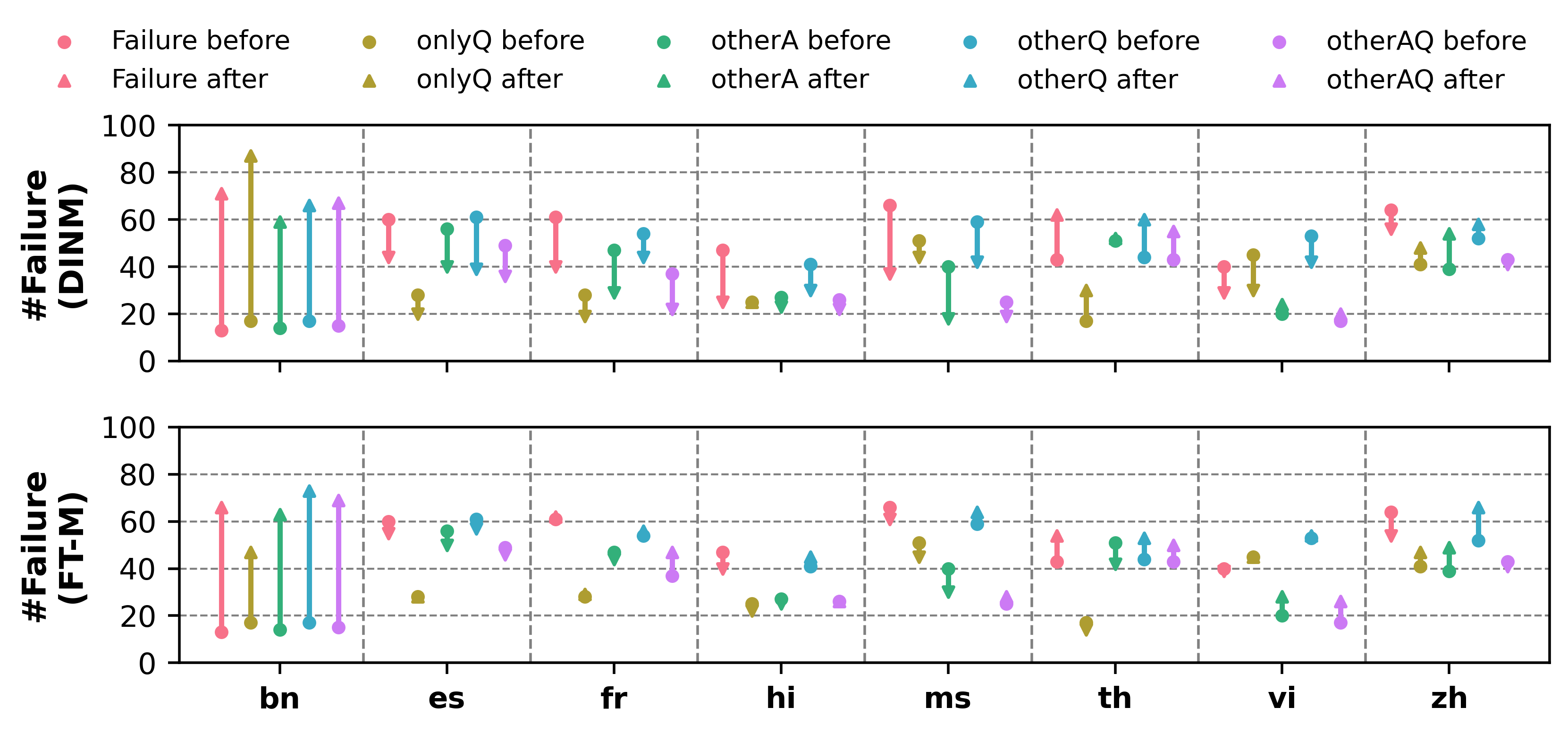}
    \caption{Number of failures and performance on OOD inputs before and after cross-lingual detoxification across languages for Llama2-7B.}
    \label{fig:cross_llama2}
\end{figure}

\begin{figure}[H]
    \centering
    \includegraphics[width=\linewidth]{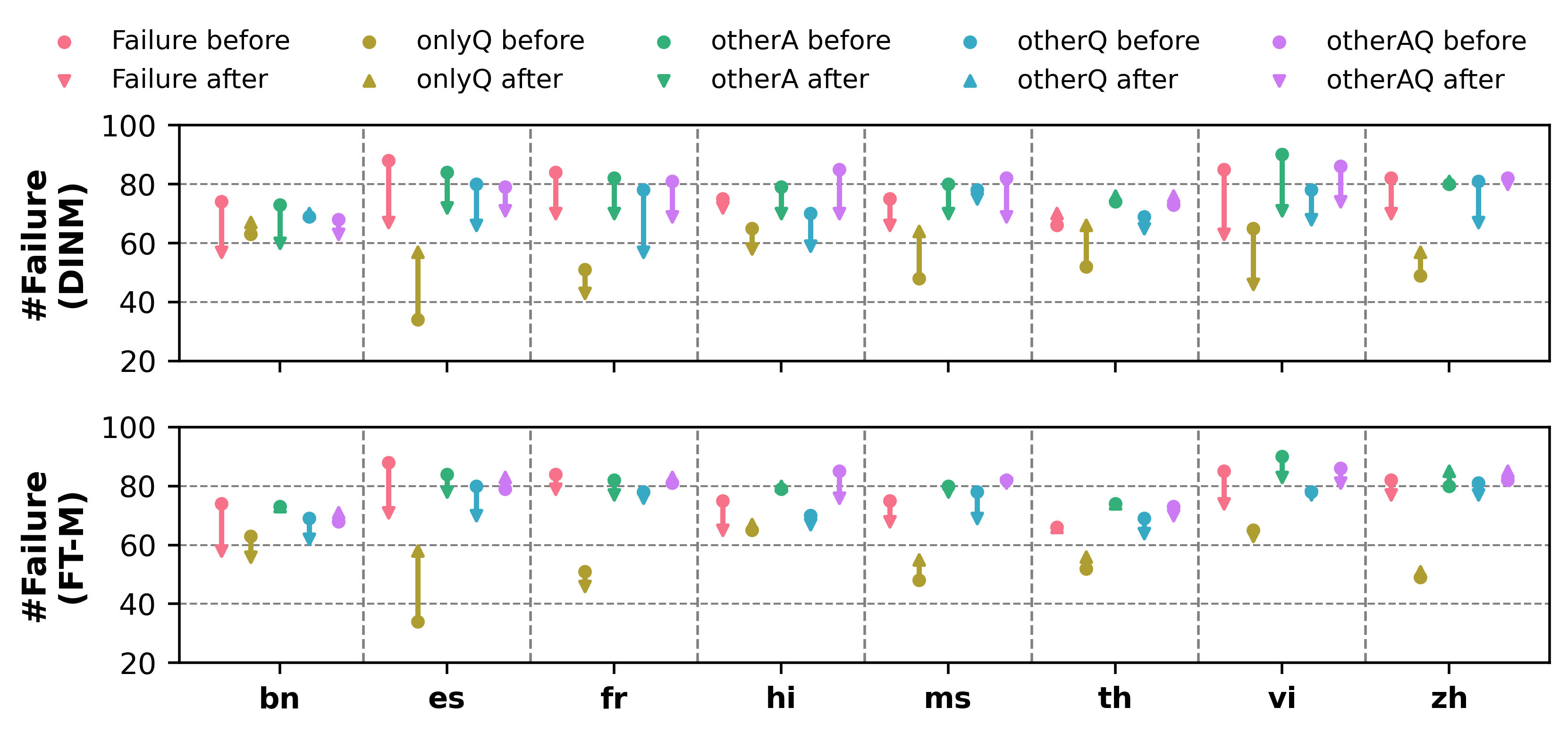}
    \caption{Number of failures and performance on OOD inputs before and after cross-lingual detoxification across languages for Llama3-8B.}
    \label{fig:cross_llama3}
\end{figure}

\begin{figure}[H]
    \centering
    \includegraphics[width=\linewidth]{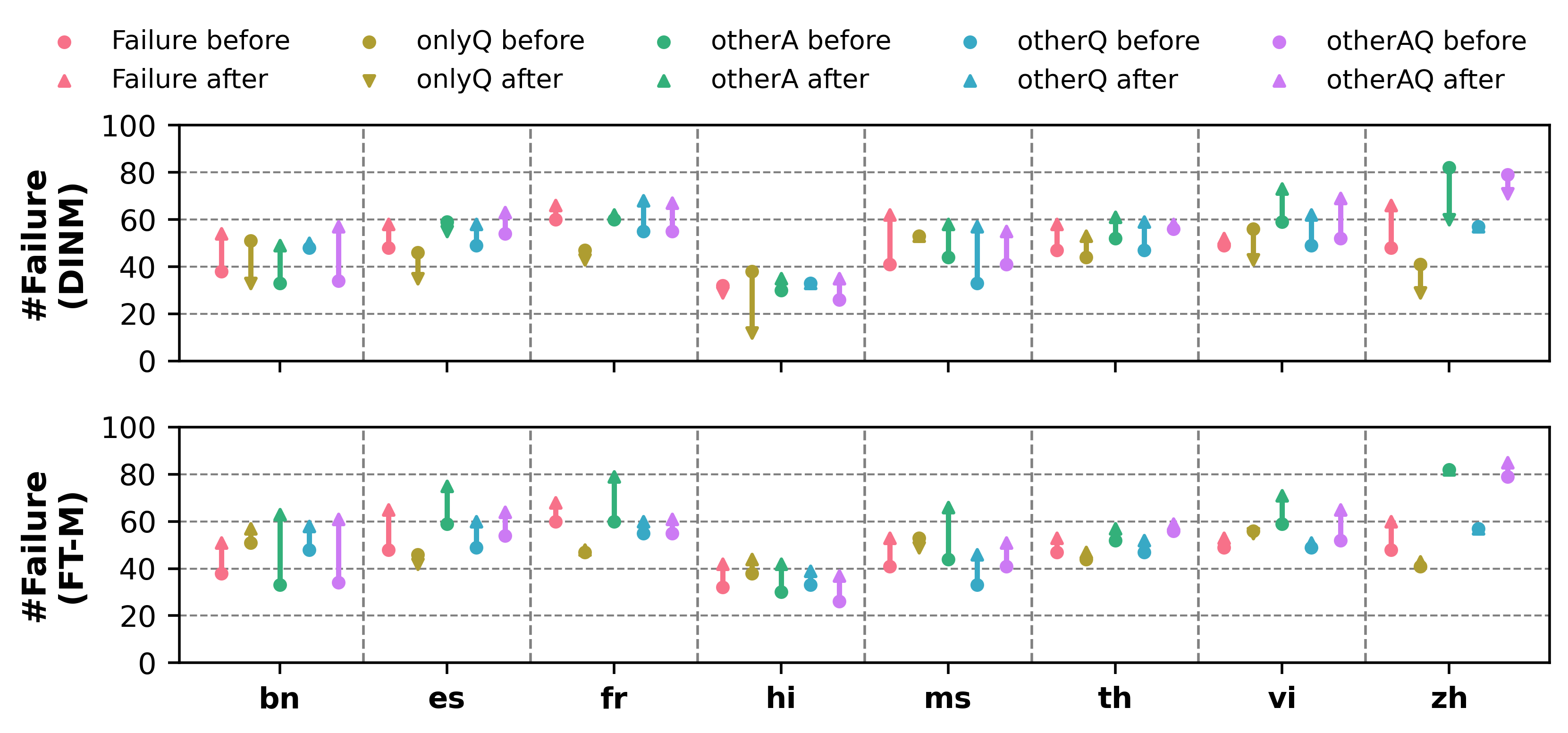}
    \caption{Number of failures and performance on OOD inputs before and after cross-lingual detoxification across languages for Ministral-8B.}
    \label{fig:cross_ministral}
\end{figure}

\begin{figure}[H]
    \centering
    \includegraphics[width=\linewidth]{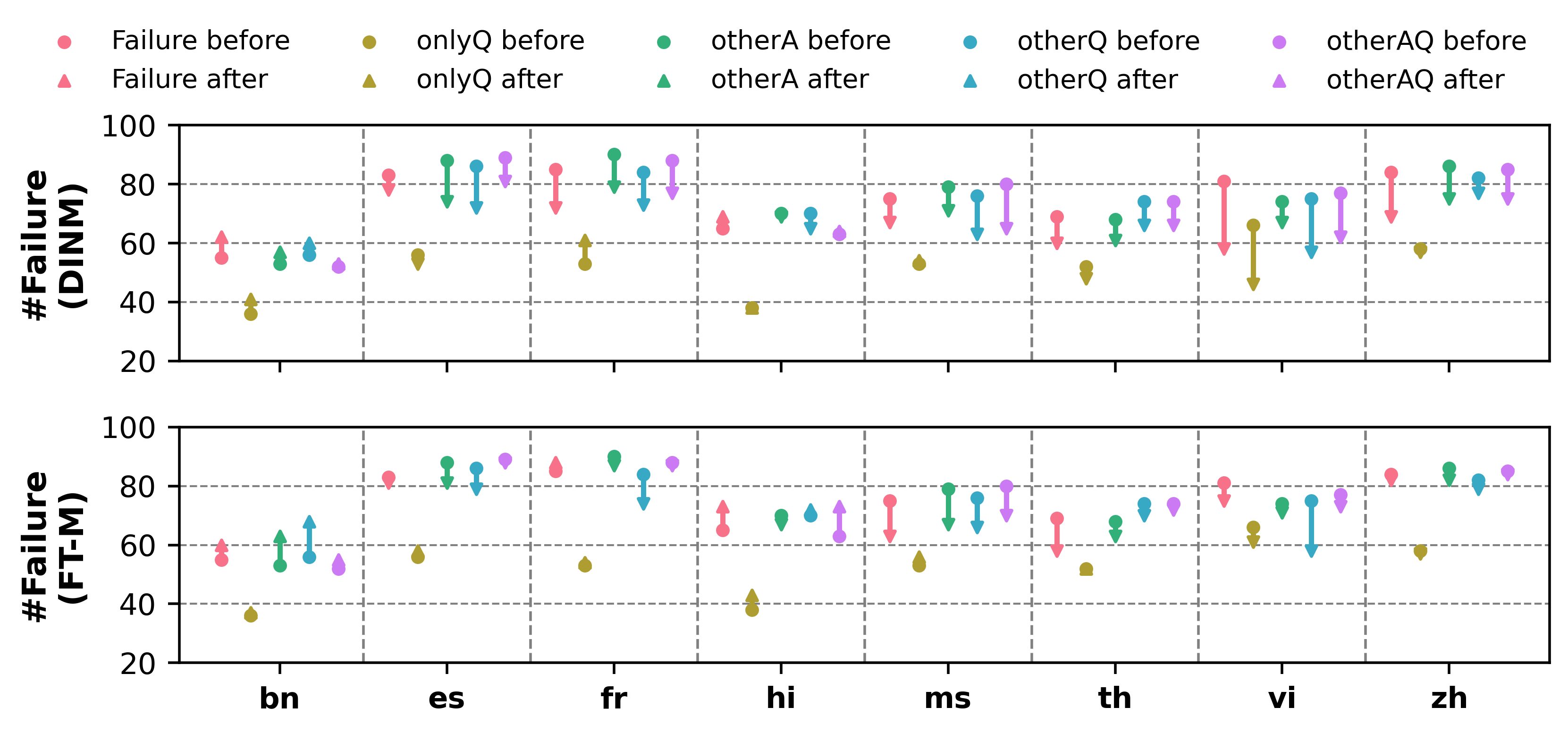}
    \caption{Number of failures and performance on OOD inputs before and after cross-lingual detoxification across languages for Mistral-7B.}
    \label{fig:cross_mistral}
\end{figure}

\begin{figure}[H]
    \centering
    \includegraphics[width=\linewidth]{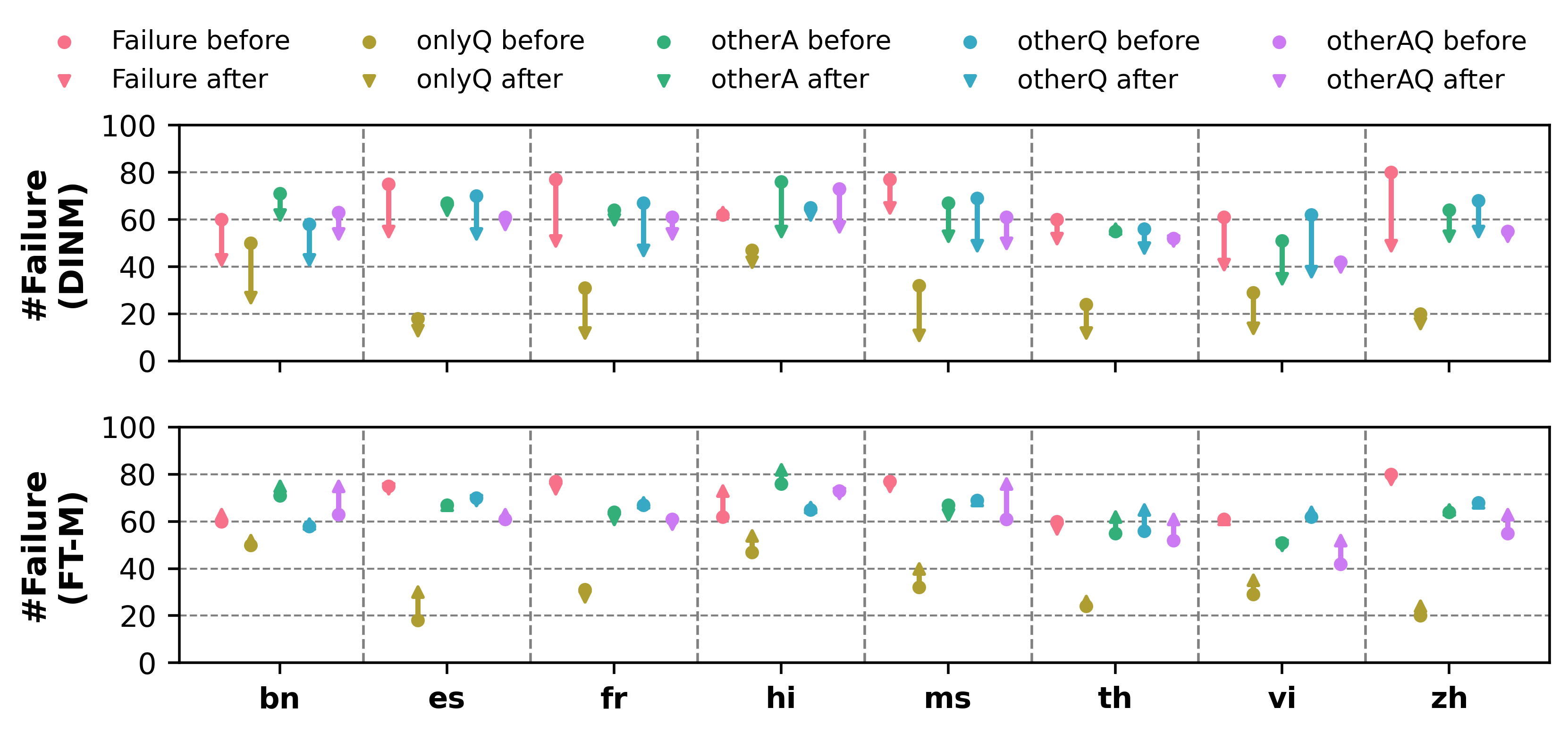}
    \caption{Number of failures and performance on OOD inputs before and after cross-lingual detoxification across languages for Qwen3-8B.}
    \label{fig:cross_qwen3}
\end{figure}

Figures~\ref{fig:cross_llama2}–\ref{fig:cross_qwen3} present the results of monolingual detoxification on Llama2-7B, Llama3-8B, Ministral-8B, Mistral-7B, and Qwen3-8B.

\end{document}